\documentclass[onecolumn]{article}         % twocolumn
\usepackage[top=3.6cm, bottom=3.2cm, left=2.5cm, right=2.5cm]{geometry}
\usepackage{tabularx}
\usepackage{graphicx}
\usepackage{multirow}
\usepackage{multicol}
\usepackage{floatrow}
\usepackage{subfigure}
\usepackage{makecell,rotating}
\usepackage{textcomp,booktabs}
\usepackage[usenames,dvipsnames]{color}
\usepackage{colortbl}
\definecolor{setgray}{gray}{.98}
\definecolor{setblue}{gray}{.58}
\newcolumntype{C}[1]{>{\centering\arraybackslash}p{#1}}
\newcolumntype{L}[1]{>{\raggedright\arraybackslash}p{#1}}
\usepackage{amssymb}
\usepackage{amsmath}
\usepackage{bm}
\usepackage{cite}
\usepackage[colorlinks=true,linkcolor=red,citecolor=blue]{hyperref}
\usepackage{url}

\begin{document}

\title{Exploiting Deep Features for Remote Sensing Image Retrieval:\\ A Systematic Investigation}
\author{Xin-Yi~Tong$^1$, Gui-Song~Xia$^{1, 2}$, Fan~Hu$^3$, Yanfei Zhong$^1$, Mihai~Datcu$^4$, Liangpei~Zhang$^1$
\vspace{3mm}
\\
$^1${\em State Key Laboratory LIESMARS, Wuhan University, China.}\\
$^2${\em School of Computer Science, Wuhan University, China.}\\
$^3${\em School of Electronic Information, Wuhan University, China.}\\
$^4${\em German Aerospace Center (DLR), Oberpfaffenhofen, Germany.}
\vspace{2mm}
\\
}
\date{}
\maketitle

\begin{abstract}
Remote sensing (RS) image retrieval is of great significant for geological information mining. Over the past two decades, a large amount of research on this task has been carried out, which mainly focuses on the following three core issues: feature extraction, similarity metric and relevance feedback. Due to the complexity and multiformity of ground objects in high-resolution remote sensing (HRRS) images, there is still room for improvement in the current retrieval approaches. In this paper, we analyze the three core issues of RS image retrieval and provide a comprehensive review on existing methods. Furthermore, for the goal to advance the state-of-the-art in HRRS image retrieval, we focus on the feature extraction issue and delve how to use powerful deep representations to address this task. We conduct systematic investigation on evaluating correlative factors that may affect the performance of deep features. By optimizing each factor, we acquire remarkable retrieval results on publicly available HRRS datasets. Finally, we explain the experimental phenomenon in detail and draw conclusions according to our analysis. Our work can serve as a guiding role for the research of content-based RS image retrieval.
\end{abstract}

\section{Introduction}
\label{sec:introduction}

With the explosive development of earth observation technologies, both the quantity and quality of remote sensing (RS) data are growing at a rapid pace \cite{Int-2002.knowledge-driven}. Millions of RS images captured by various satellite sensors have been stored in massive archives \cite{Int-2003.information.I,Int-2005.information.II}. To make full use of big RS data, efficient information management, mining and interpretation methods are urgently needed. During the past decades, significant efforts have been made in developing accurate and efficient retrieval methods to search data of interest from large RS archives \cite{Int-2013.a.review,Int-2007.Introduction,Int-2010.Introduction,Int-2017.PatternNet}.

Primal RS image retrieval systems generally used geographical area, time of acquisition or sensor type as queries \cite{Int-1993.the.intelligent,Int-1997.titan}. These approaches might be very imprecise and inefficient because text-based image retrieval rely largely on manually annotated keywords \cite{Int-1998.GeoBrowse}, which are less relevant to the visual content of RS images. As content-based image retrieval \cite{Int-1994.efficient,Int-1995.content-based.image} was proposed in the early 1990s, the performance of RS image retrieval approaches has been remarkably improved. New architectures for RS image archives were constructed, where RS images were stored \cite{Int-1997.new} and retrieved \cite{Int-1998.query,Int-2000.Image} based on visual content. So far, several mature RS retrieval systems have come into service \cite{Int-1998.GeoBrowse,Int-2002.VisiMine,Vis-2005.large-scale,Vis-2006.automated,
Vis-2006.a.framework,Vis-2007.geoiris,Int-2002.knowledge-driven,Int-2003.information.I,Int-2005.information.II,Int-2010.System.design}.

Content-based image retrieval takes images as queries, rather than keywords, whose performance therefore is extremely dependent on the visual features \cite{CBIR1,CBIR3}. For promoting the accuracy of RS image retrieval, early studies mainly focused on seeking various feature representation methods, hoping to find more discriminating image features \cite{Int-1995.content-based.retrieval,Int-1995.system,Int-1996.retrieving,Int-1998.meta,Int-1998.spatial.I,Int-1998.spatial.II,
Vis-2000.Gibbs,Vis-2002.a.retrieval.system,Vis-2002.retrieval.of.remotely,Vis-2004.remote.sensing.imagery,Vis-2004.using.texture,
Vis-2005.local.shape,Vis-2007.camparing,Vis-1999.An.environment,Vis-2001.Query-by-shape,Vis-2005.Wavelet.features,Vis-2007.A.systematic,
Vis-2014.Improved.color,Vis-2015.Multiscale} or feature combinations \cite{Int-1999.content-based,Vis-1997.deriving,Vis-2000.multi-level,Vis-2004.integrated,Vis-2004.retrieval.using.texture,
Vis-2004.semantics-based,Vis-2009.retrieval,Vis-2009.searching,Vis-2009.Prototype.system,Vis-2014.Remote.Sensing,Vis-2015.Dual-tree}. Nevertheless, due to the drastically increasing volume and complexity of RS data, visual features may become subjective and ambiguous in some situations \cite{Int-2013.a.review}. Consequently, the performance of the basic RS retrieval systems was no longer satisfactory. To solve this problem, on the one hand, researchers proposed to select or design the most suitable similarity metric for some specific tasks \cite{Sim-2004.comparative,Sim-2008.A.similarity,Sim-2012.an.endmenber,Sim-2012.dictionary}, which can adaptively amend the degree of similarity between image feature vectors. On the other hand, researchers applied relevance feedback to RS retrieval system \cite{Int-2000.interactive,Int-2002.knowledge-driven,Int-2003.information.I,Int-2005.information.II,Rel-2007.interactive,Rel-2015.a.novel,
Rel-2001.Fast.retrieval,Rel-2002.Probabilistic,Rel-2004.Interactive,Rel-2006.Scalable,Rel-2007.Learning-unlearning,Rel-2007.Semantic-sensitive,
Rel-2010.Visual.information,Rel-2010.Visual-semantic,Rel-2011.An.interactive}, aiming to capture the exact intentions of the users and return retrieval results that meet user demand. As described above, feature extraction, similarity metric and relevance feedback constitute the three core issues of modern RS image retrieval framework.

In this paper, the main issue we focus on is the visual features. According to the approaches of feature extraction, the existing retrieval methods can be divided into three categories: methods based on low-level features, methods based on mid-level features and methods based on high-level features. Low-level features are always designed by human on the basis of engineering skills and domain expertise. Diverse low-level features have been exploited for RS retrieval, mainly including spectral features \cite{Int-1995.content-based.retrieval,Int-1995.system,Int-1996.retrieving,Vis-2002.a.retrieval.system,Vis-2002.retrieval.of.remotely}, texture features \cite{Vis-2000.Gibbs,Vis-2004.remote.sensing.imagery,Vis-2004.using.texture,Vis-2004.retrieval.using.texture,Vis-1997.deriving,Vis-2014.morphological,
Vis-2005.Wavelet.features,Vis-2007.A.systematic,Vis-2014.Improved.color,Vis-2015.Multiscale} and shape features \cite{Vis-2005.local.shape,Vis-2011.entropy-balanced,Vis-1999.An.environment,Vis-2001.Query-by-shape}. In contrast, mid-level features can represent more discriminating information by encoding raw features using bag-of-words (BoW) \cite{Mid-BoW}, Fisher vector (FV) \cite{Hig-FV}, vector locally aggregated descriptors (VLAD) \cite{Mid-VLAD} or their variants. However, owing to the changes of photography scale, orientation and illumination, which makes the relevant ground objects have quite different appearance, the above hand-crafted features lose their effectiveness of discriminating high resolution remote sensing (HRRS) images.

To overcome this difficulty, researchers made use of high-level features derived from Convolutional Neural Networks (CNNs) for HRRS image retrieval. In recent literatures \cite{Hig-2016.learning,Hig-2017.visual,Hig-2017.Retrieving.Aerial}, CNN features have proved to be of strong discrimination ability and able to dramatically improve the retrieval performance. CNNs are deep hierarchical architectures with parameters (or weights) of each layer learned from large-scale datasets \cite{AlexNet}. The pre-trained CNNs can be well transferred to relatively small datasets for feature learning \cite{appSum1,appSum4}. Nevertheless, when a CNN model trained for classification is used for domain-specific retrieval, its transferability and adaptability to the target data is likely to be unreliable. Various factors involved with transferability may limit the performance of deep feature-based retrieval methods. With this in mind, we intend to further investigate how to better use deep features for content-based HRRS image retrieval task.

We first analyze the retrieval framework and present a comprehensive review on the following three core issues: feature extraction, similarity metric and relevance feedback, so as to complement existing surveys in literatures \cite{Int-2013.a.review,Int-2007.Introduction,Int-2010.Introduction,Int-2017.PatternNet}. Then, we focus on the feature extraction issue and delve into deep features to fully advance the state-of-the-art in HRRS image retrieval. We investigate almost all influencing factors concerned to the property of deep features, including CNN architectures, depth of layers, aggregation method for feature maps, dimension of features and fine-tuning. In addition, we propose multi-scale concatenation and multi-patch pooling methods to further promote the retrieval performance.

In summary, this paper mainly contributes in the following aspects:
\begin{itemize}
\item[-] We provide a comprehensive review of content-based RS image retrieval, covering the three key issues: feature extraction, similarity metric and relevance feedback.
\item[-] We investigate systemically how to utilize deep features for HRRS image retrieval. Comprehensive influencing factors are assessed, and noteworthy experimental results are achieved on three public HRRS datasets.
\item[-] We analyze the experimental phenomena in detail, some of which are generalized, and some are data-dependent. We draw many instructive conclusions from thorough analysis, which can play a guiding role for domain-specific retrieval problems.
\end{itemize}

\section{Review on RS Image Retrieval}
\label{sec:review}
In this section, we firstly make a detailed introduction of content-based RS image retrieval, and then comprehensively review the existing research in this field. We take three key aspects, feature extraction, similarity metric and relevance feedback, into consideration and analyze their role in the retrieval task.

\begin{figure}[!tbp]
\centering
\includegraphics[width=0.65\linewidth]
{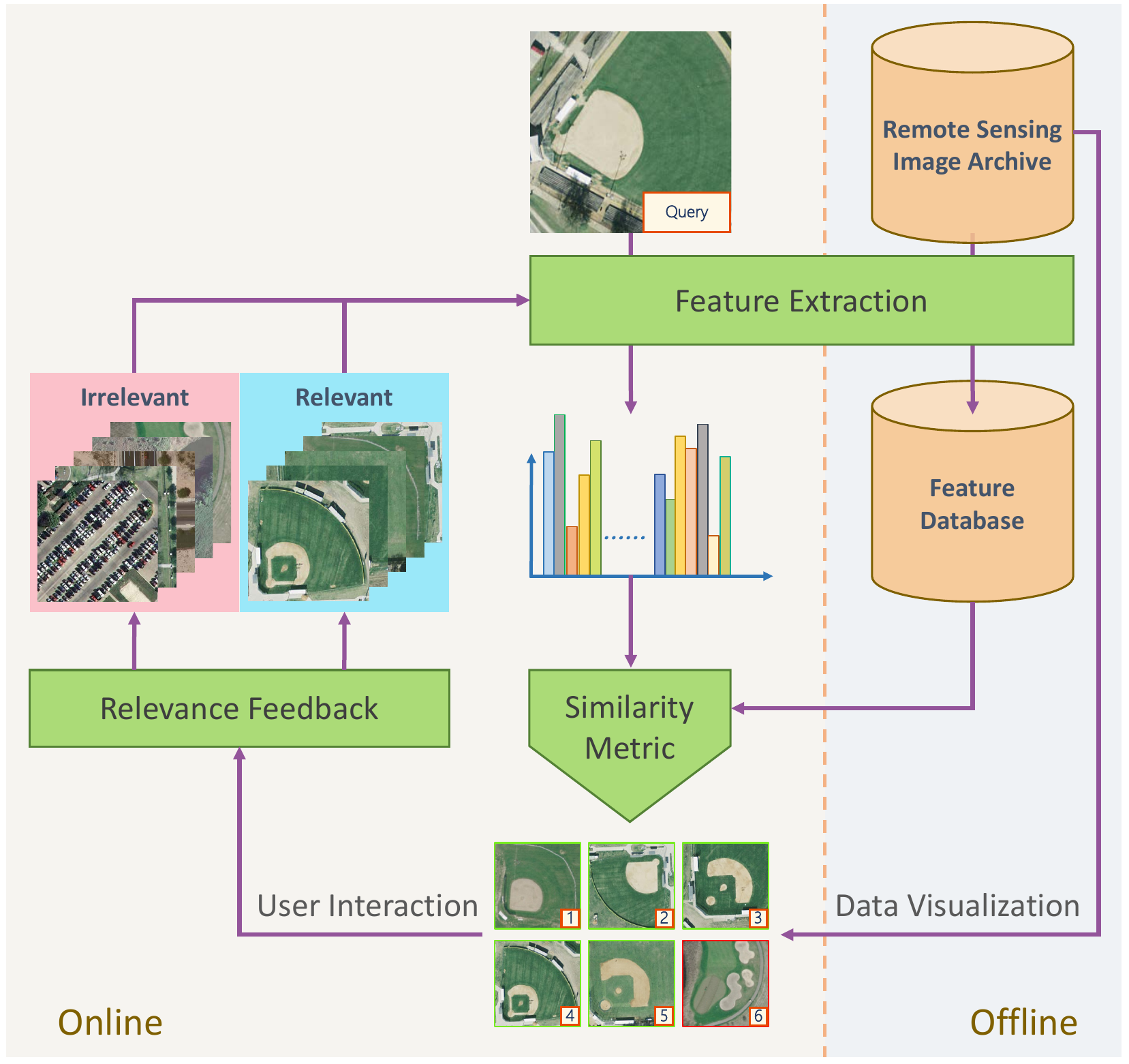}
\caption{The framework of content-based RS image retrieval system.}
\label{figure:FlowChart0}
\end{figure}

\subsection{An Overview of RS Image Retrieval}
The goal of content-based RS image retrieval is to find a set of images that contain the content desired by the users from RS archives. We indicate the images stored in database to be retrieved as reference images. If the retrieval system returns reference images containing relevant visual content, we regard them as correct retrieval results.

A content-based image retrieval framework at least consists of two stages \cite{CBIR1,CBIR2}. The first stage extracts image features to describe the physical object and scene of both the input query image and reference images. The second stage calculates the visual similarity between the query image and each reference image based on feature vectors, and then returns a ranked list of relevant images ordered by the degree of similarity. Moreover, if visual features and similarity metrics have limited ability to accurately measure the relationship between image contents, relevance feedback can be used to interactively revise the initial ranking \cite{Rel-1996.Bayesian,Rel-1998.mindreader}. The overall framework of a RS image retrieval system is illustrated in Fig. \ref{figure:FlowChart0}.

To gain an insight into content-based image retrieval of RS imagery, we present a comprehensive review focusing on the above three core issues. Though there has been a few surveys on broad content-based RS image retrieval research \cite{Int-2013.a.review,Int-2007.Introduction,Int-2010.Introduction,Int-2017.PatternNet}, they particularly give attention to some single aspects of RS data mining. Our work can serve as a thorough complement for the previous surveys.

\subsection{How to Represent RS Images}
RS image retrieval methods can be divided into three categories based on the way of feature extraction: methods based on low-level features, methods based on mid-level features and methods based on high-level features. We introduce the relative literatures at length in the following.

\subsubsection{Methods based on low-level features}
Visual content can be typically defined by a set of hand-crafted features that describe the spectral, texture or shape information of RS images.

Spectral features are one of the simplest features, yet they describe the most prominent information of RS images \cite{Oth-2005.Study.on}. Spectral features have been utilized for various RS retrieval works \cite{Int-1995.content-based.retrieval,Int-1995.system,Int-1996.retrieving,Vis-2002.a.retrieval.system,Vis-2002.retrieval.of.remotely}. They encode the reflectance of the corresponding areas of the Earth¡¯s surface, resulting in serious sensitivity to noise and illumination change.

Texture features are generally understood as ordered structures composed of a group of pixels \cite{Vis-GLCM}. A number of texture features have been applied to RS image retrieval in the form of single feature \cite{Vis-2000.Gibbs,Vis-2004.remote.sensing.imagery,Vis-2004.using.texture,Vis-2005.Wavelet.features,Vis-2007.A.systematic,Vis-2014.Improved.color,Vis-2015.Multiscale} or combination of multiple features \cite{Vis-2004.retrieval.using.texture,Vis-1997.deriving,Vis-2014.morphological}. Commonly used texture features include gray level co-occurrence matrices (GLCM) \cite{Vis-GLCM}, wavelets \cite{Vis-wavelet,Vis-wavelet.based.histogram}, Gabor filters \cite{Vis-Gabor,Vis-Gabor.coefficient} and local binary patterns (LBP) \cite{Vis-LBP}. However, they do not fully reflect the essential features of objects because texture is only a characteristic of surface.

Shape features are important cue for content recognition of RS images \cite{Vis-1999.An.environment,Vis-2001.Query-by-shape,Vis-2005.local.shape,Vis-2011.entropy-balanced}. They have been used for infrared image retrieval \cite{Vis-2005.local.shape} and object retrieval in optimal image \cite{Vis-2011.entropy-balanced}. Shape features describe the outline or area information of ground objects, but have little ability to capture their spatial relationship information.

Other types of features are also proposed for RS image retrieval. Scale-invariant feature transform (SIFT) \cite{Vis-SIFT} has been proved to be more effective than texture features in scene retrieval \cite{Vis-2007.camparing}. Structural features derived from shape ensembles and relationships \cite{Vis-topographic,Vis-indexing} also provide satisfactory performance \cite{Vis-2010.structural,xia2014texture,xia2017SCOPE}. In addition, researchers have explored combinations of diverse low-level features to improve the retrieval results \cite{Int-1999.content-based,Vis-2000.multi-level,Vis-2004.integrated,Vis-2004.semantics-based,
Vis-2005.large-scale,Vis-2006.automated,Vis-2006.a.framework,Vis-2007.geoiris,Vis-2009.retrieval,
Vis-2009.searching,Vis-2013.scene.semantic.matching,Vis-2009.Prototype.system,Vis-2014.Remote.Sensing,Vis-2015.Dual-tree}. Different visual features make up for each other¡¯s defects, hence their combinations have stronger discriminating ability.

\subsubsection{Methods based on mid-level features}
In contrast with low-level features, mid-level features embed raw descriptors into visual vocabulary space and encode feature spatial distribution to capture semantic concepts. Mid-level features are more invariant to changes of scale, rotation or illumination, and they can better represent the complex textures and structures with more compact feature vectors. The general pipeline to extract mid-level features is firstly obtaining local image descriptors, such as spectral, texture or local invariant features, and then aggregating them into holistic representations using encode methods, e.g., BoW \cite{Mid-BoW}, FV \cite{Hig-FV}, and VLAD \cite{Mid-VLAD}.

BoW \cite{Mid-BoW} is a widely used basic encoding method, it employs k-means clustering to construct visual codebook and counts local features into the histogram of codebook. It has been utilized in some RS image retrieval research and has achieved desired results. Concretely, \cite{Mid-2013.geographic,Mid-2014.bag,Mid-2015.An.improved} have shown the effectiveness of encoded features compared with local low-level features.

VLAD \cite{Mid-VLAD} is an advanced version of BoW, apart from feature distribution, it additionally counts the distance between local features and cluster centers. VLAD is applied to encode local pattern spectra \cite{Mid-pattern} and obtains high-precision retrieval results on HRRS images \cite{Mid-2016.retrieval}. In \cite{Mid-2014.performance}, the experimental results demonstrate that BoW behaves better in calculation speed while VLAD behaves better in indexing accuracy.

Multi-scale spatial information has also been exploited for feature encoding. For instance, spatial pyramid matching based on sparse codes (ScSPM) \cite{Rel-ScSPM} fuses holistic and local features to enhance the discrimination of mid-level features \cite{Rel-2016.a.three-layered}. Except for the above methods, other unsupervised feature learning methods have also been employed to construct features with higher level of semantic information. Such as auto-encoder \cite{Mid-2015.High-resolution} and hierarchical neural networks \cite{Mid-2016.Content-Based}.

\subsubsection{Methods based on high-level features}
The hierarchical architecture of CNN models can simulate very complex nonlinear functions and automatically learn parameters during the training process \cite{AlexNet}. Therefore, CNN models are able to capture the essential characteristics of training data so as to represent discriminating visual features \cite{fea2}.

Some RS image retrieval works based on high-level features have been represented up till now. The approaches include obtaining features by existing CNNs from convolutional (conv.) layers or fully-connected (FC) layers \cite{Hig-2017.visual}, fine-tuning off-the-shelf CNN models with domain-relevant datasets \cite{Hig-2017.Retrieving.Aerial}, or developing tailored CNN architectures \cite{Hig-2016.learning} and training it with large scale RS dataset \cite{Hig-AID}, etc.

Nevertheless, there is no a comprehensive investigation on deep feature-based HRRS image retrieval. Besides, whether the research conclusions of specific retrieval contexts are transferable to other situations is unknown, since the diversity of data domain has some degree of impact on the feature description. Therefore, how to optimize the performance of deep feature-based HRRS image retrieval is still a problem need to be solved.

\subsection{How to Measure Feature Similarity}
Similarity metric (or distance function) is a function that defines the distance between visual feature vectors \cite{Sim-2008.A.similarity}, which is one of the foundations of pattern recognition.
In a RS image retrieval task, different similarity metric may lead to different ranking results. In \cite{Sim-2004.comparative}, eight similarity metrics are investigated. Similarity metrics examined in this work can be divided into two major categories: general feature vector-based metrics and histogram vector-based measures. This work intuitively demonstrates the importance of similarity metrics in retrieval process.

Apart from selecting the appropriate similarity metrics, distance functions can also be manually constructed for specific retrieval situation. For instance, in \cite{Sim-2008.A.similarity}, an informational similarity metric is introduced for compressed RS data mining. In \cite{Sim-2012.an.endmenber}, a hyper-spectral image distance is developed. In \cite{Sim-2012.dictionary}, dictionary-based similarity metrics are employed for retrieval in different hyper-spectral image datasets, demonstrating the applicability of dictionary-based similarity metrics for hyperspectral image retrieval.

However, manually constructing a similarity metric may be inefficiency and not robust to different data source, metric learning can be an ideal alternative. In contrast to hand-crafted similarity metrics, metric learning is capable of automatically learning distance function for a specific retrieval situation according to task requirement \cite{Sim-metric.learning.I,Sim-metric.learning.II,Sim-metric.learning.III,Sim-metric.learning.IV}. Unsupervised metric learning has been successfully applied to RS retrieval, for example, \cite{Sim-2016.Region-Based} models RS images with graphs and uses an unsupervised graph-theoretic method to measure the similarity between the query graph and the graphs of images in the archive. Besides, deep learning-based metric learning approaches have been investigated. In \cite{Sim-2015.NetVLAD}, geographic coordinates are treated as weakly supervised information and used to train a triplet network for street view image retrieval.

\subsection{How to Optimize Ranking Result}
In the case where the visual features are discriminating, and the similarity metric is adaptive, the ranking results of content-based RS image retrieval may still be unsatisfactory \cite{Rel-2015.a.novel}. The intelligent feedback techniques therefore become essential for RS retrieval systems.

Relevance feedback can iteratively optimize the retrieval results according to the previous ranking. Once the ranking of the initial retrieval is returned, there are two ways to select a subset of relevant images: automatically selection and manually selection, which are applied to pseudo relevance feedback and explicit relevance feedback respectively \cite{Rel-automatic.manual}.

In pseudo relevance feedback \cite{Rel-automatic.manual}, the top several returned results are regarded as relevant images, and their features are used for query expansion \cite{Rel-2016.a.three-layered, Rel-SimpleMKL}. Then the fused feature vector is considered as a new query and able to generate more exact ranking list.

In contrast, in explicit relevance feedback \cite{Rel-automatic.manual}, the retrieved images are marked as ¡°relevant¡± or ¡°irrelevant¡± manually by the users at every feedback round. There are three different methods to re-estimate the target query, namely query-point movement and re-weighting method \cite{Rel-1995.optimization,Rel-1998.mindreader}, probability distribution-based method \cite{Rel-1996.Bayesian,Rel-2000.the.Bayesian} and machine learning-based method \cite{Rel-2001.support,Rel-2001.a.neural}.

The idea of query-point movement and re-weighting method is to adjust the query point in the feature space according to the users¡¯ feedback, and then use the adjusted query point to re-calculate the ranking list \cite{Rel-2001.Fast.retrieval}.

The probability distribution-based method aims to minimize the probability of retrieving irrelevant images. Specifically, assume there is a mapping from the visual features to the image categories, and the purpose of probability distribution-based relevance feedback is to find the optimal mapping that can minimize the error probability \cite{Rel-2002.Probabilistic,Rel-2011.An.interactive,Int-2000.Image,Int-2000.interactive,Int-2002.knowledge-driven}.

Machine learning-based relevance feedback can be considered as a binary-classification problem: the relevant retrieved images are positives while the irrelevant retrieved images are negatives \cite{Rel-2015.a.novel}. In each iteration, the classifier can be trained with the feedback samples of the current round, or with the combination of the current and the former feedback samples via incremental learning. It returns image ranking according to the category scores derived from the classifier. Commonly used classifiers include decision tree \cite{Rel-2004.Interactive,Rel-2010.Visual-semantic}, Bayesian networks \cite{Rel-2007.Semantic-sensitive}, support vector machine (SVM) \cite{Rel-2006.Scalable,Rel-2007.Learning-unlearning,Rel-2007.interactive,Rel-2015.a.novel} and so on.

Apart from the above literatures, there are also many works aiming at improving retrieval efficiency for RS images, including taking advantages of distributed computation \cite{Oth-2015.distributed}, applying tree structures \cite{Vis-2005.large-scale,Vis-2006.automated,Vis-2006.a.framework,Vis-2007.geoiris,Vis-2009.searching,Vis-2011.entropy-balanced} and utilizing hash codes \cite{Vis-2005.local.shape,Vis-polygon.I,Vis-polygon.II,Oth-2014.kernel-based,Oth-2016.hashing-based}.

\begin{figure}[htb!]
\centering
\includegraphics[width=0.65\linewidth]
{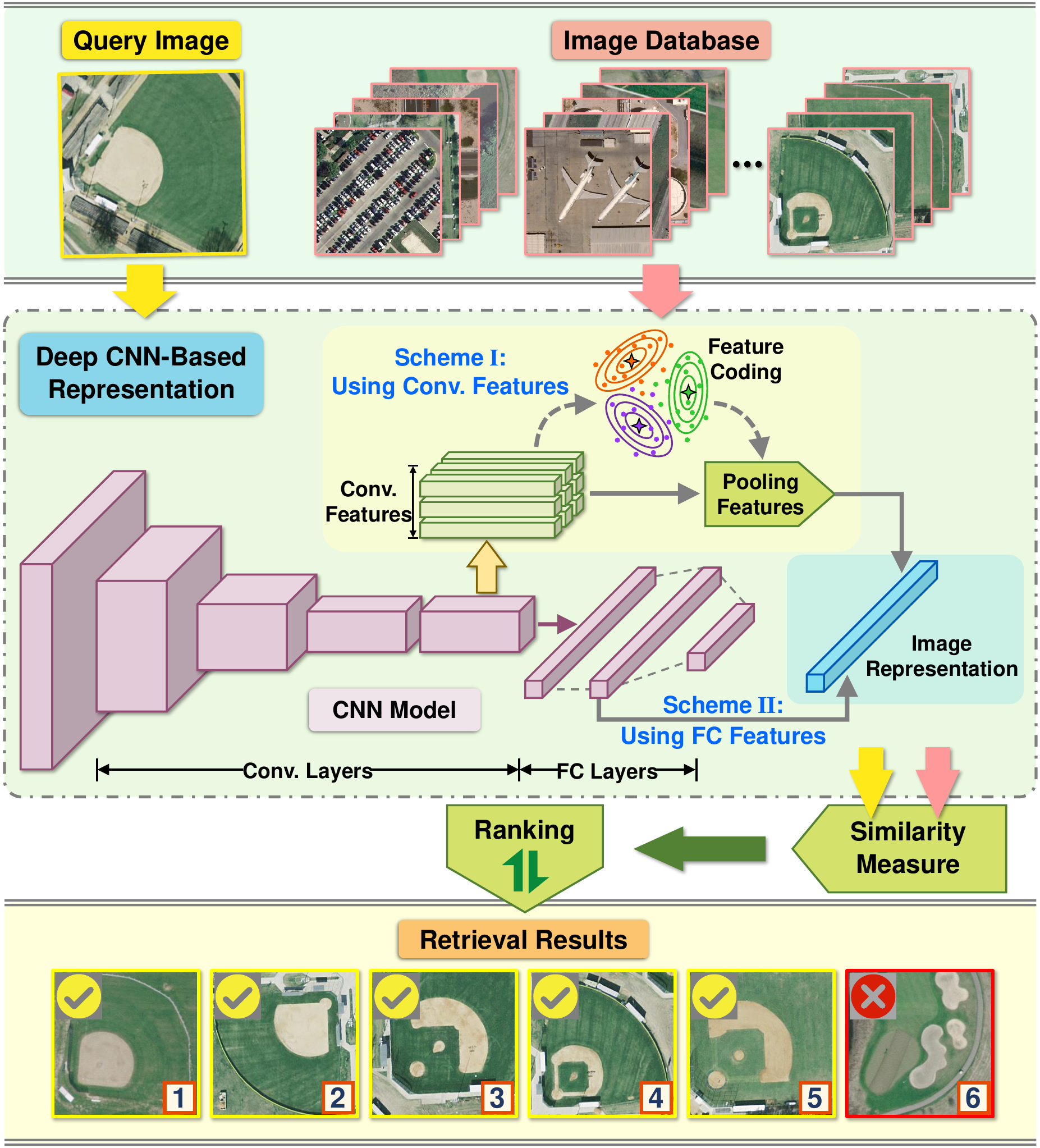}
\caption{The pipelines of scheme (I) and scheme (II). }
\label{figure:FlowChart1}
\end{figure}

\section{Deep Features for RS Image Retrieval}
\label{sec:approaches}
Although some literatures have made advantages of deep features for RS retrieval task, there still no comprehensive research on how to optimize the transferability of CNN models to RS retrieval. With this in mind, we investigate almost all variables concerned to the property of deep features on several public HRRS datasets and analyze the effects of each variable.

An elementary content-based image retrieval framework is at least composed of two stages. For an image dataset which contains $N$ images, the first stage extracts the visual features $\bm{f}^q$ and $\bm{f}^r$ respectively from the query image $\bm{I}^q$ and all the reference images $\bm{I}^r$, for $r=1,...,N$. The second stage calculates the distances $D_{qr}=d(\bm{f}^q,\bm{f}^r)$ between extracted feature vectors and then ranks retrieved images according to the values of $D_{qr}$, i.e., the more similar $\bm{f}^q$ and $\bm{f}^r$ are, the lower $\bm{I}^r$ ranks, where $d(\cdot,\cdot)$ stands for a distance function.

Because conv. and FC features derived from different depths in CNN architecture, they are at different representation levels. Conv. features correspond to local responses of every image region, while FC features contain global information of the holistic image, this diversity may lead to different retrieval performance. Moreover, the off-the-shelf CNN models may have limited capability of transferring to RS domain, which is likely to degrade the performance of HRRS image retrieval.

On account of the aforementioned considerations, we make use of five representative CNN models: CaffeNet \cite{CaffeNet}, VGG-M \cite{VGGNets}, VGG-VD16 \cite{VGG-VDNets}, VGG-VD19 \cite{VGG-VDNets}, GoogLeNet \cite{GoogLeNet}, and exploit various approaches for feature extraction.

\begin{figure}[t]
\includegraphics[width=0.8\linewidth]
{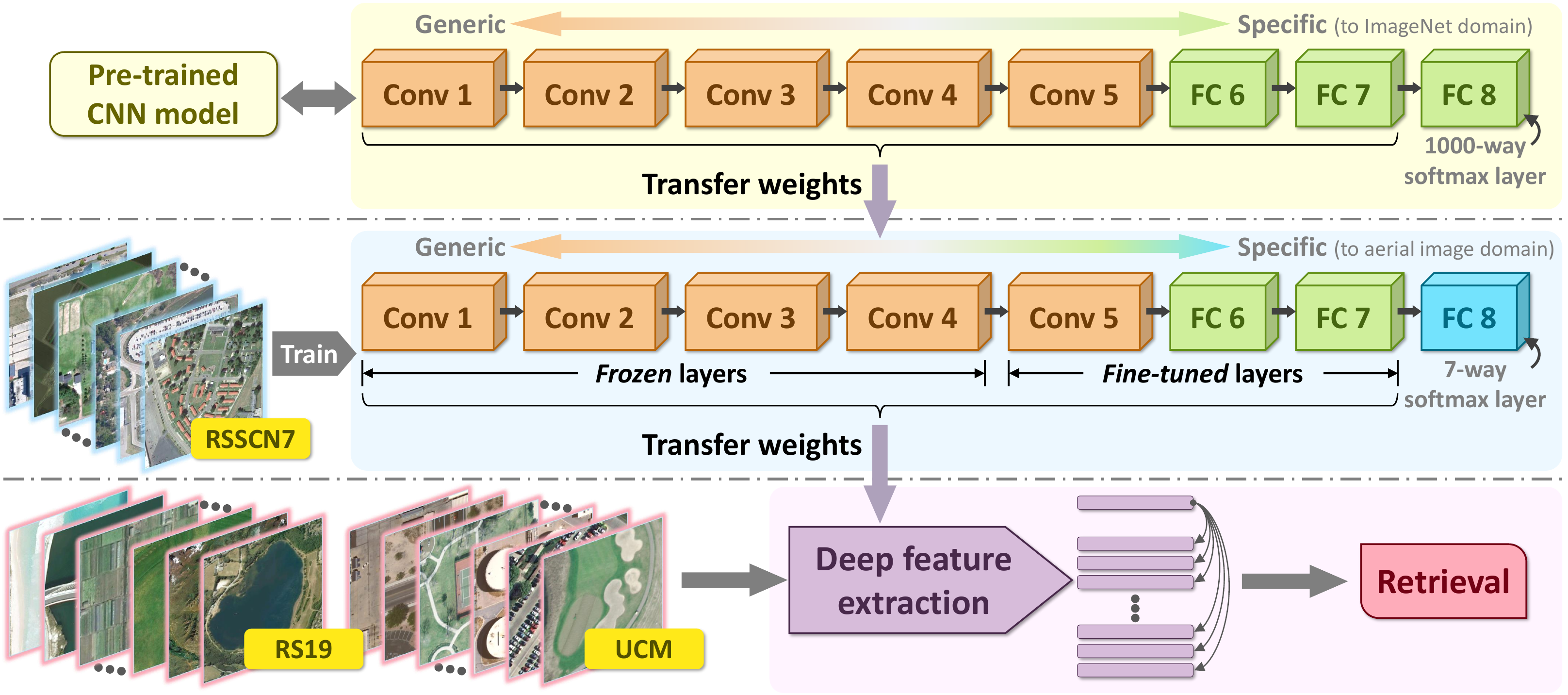}
\caption{Overview of fine-tuning for scheme (III).}
\label{figure:FlowChart2}
\end{figure}

We mainly conduct three schemes: extracting deep features from conv. and FC layers of the pre-trained CNNs respectively, the process of which is illustrated in Fig. \ref{figure:FlowChart1}; and fine-tuning the off-the-shelf CNN models for targeted feature extraction, demonstrated in Fig. \ref{figure:FlowChart2}.

\subsection{Scheme (I): Employing Conv. Features}
\subsubsection{Convolutional Features}
When passing an image $\bm{I}$ through a CNN, the outputs from conv. layers are feature maps, in which each element corresponds to a receptive field of the input image. Suppose the responses of a certain conv. layer form $L$ feature maps and the size of each feature map is $W\times H$. The activations of this layer can be interpreted as $W\times H$ $L$-dimensional vectors, where the channel number $L$ depends on the inherent structure of CNN, and the spatial resolution $W\times H$ depends on the architecture of CNN, the adopted layer, and the size of image $\bm{I}$.

The similarity metric between images is calculated based on their feature vectors, hence it is necessary to aggregate the 3-dimensional feature maps into 2-dimentional feature vectors. We apply several promising aggregation strategies to process feature maps, which can be divided into pooling and encoding methods.

\subsubsection{Pooling Methods}
\label{pooling}
We utilize five pooling methods for conv. feature aggregation: max pooling, mean pooling, hybrid pooling \cite{Hybrid}, sum-pooled convolutional features (SPoC) \cite{SPoC}, and cross-dimensional weighting and pooling features (CroW) \cite{CroW}. When dealing with feature maps using pooling methods, we treat conv. activations as $L$ two-dimensional matrices and each of them composed of $W\times H$ elements: $a_{1,1}^{z}$,$a_{1,2}^{z}$,...,$a_{W,H}^{z}$.
\begin{itemize}
\item[-]\emph{Max Pooling \cite{Hybrid}}:
Max pooling generates $L$-dimensional feature vector $\bm{f}=[f_{1},\cdots,f_{z},\cdots,f_{L}]\in\mathbb{R}^{L}$, where every element in resulting representation is simply the max activation of a corresponding feature map:
\begin{equation}
f_{z}=\max\limits_{1\leq x\leq W,1\leq y\leq H}a_{x,y}^{z}.
\end{equation}
\item[-]\emph{Mean Pooling \cite{Hybrid}}:
Analogous, the $L$-dimensional output $\bm{f}=[f_{1},\cdots,f_{z},\cdots,f_{L}]\in\mathbb{R}^{L}$ of mean pooling is a set of average values yielded from corresponding feature maps:
\begin{equation}
f_{z}=\frac{\sum_{y=1}^{H}\sum_{x=1}^{W}a_{x,y}^{z}}{WH}.
\end{equation}
\item[-]\emph{Hybrid Pooling \cite{Hybrid}}:
The feature vector produced by hybrid pooling is the intuitional concatenation of max pooling and mean pooling representations, therefore the hybrid pooling representation is with a dimension of $2L$.
\item[-]\emph{SPoC \cite{SPoC}}:
SPoC representation is acquired with center-prior Gaussian weighting on spatial of feature maps as $\bm{f}=[f_{1},\cdots,f_{z},\cdots,f_{L}]\in\mathbb{R}^{L}$, followed by sum pooling:
\begin{equation}
f_{z}=\sum_{y=1}^{H}\sum_{x=1}^{W}g_{(x,y)}a_{x,y}^{z}.
\end{equation}
The function $g_{(x,y)}$ is constructed based on Gaussian weighting scheme:
\begin{equation}
g_{(x,y)}=\exp\bigg\{-\frac{(y-\frac{H}{2})^2+(x-\frac{W}{2})^2}{2\sigma^2}\bigg\},
\end{equation}
where $\sigma$ is set to be $1/3$ of the distance between the center and the closest boundary of the input image.
\item[-]\emph{CroW \cite{CroW}}:
CroW is a promotion version of SPoC with specific non-parametric schemes for both spatial and channel wise weighting. Firstly, activations with positive values of each feature map channel are counted as:
\begin{equation}
\phi^{z}=\frac{\sum_{y=1}^{H}\sum_{x=1}^{W}\theta_{x,y}^{z}}{WH},
\end{equation}
where $\theta_{x,y}^{z}=1$ if $a_{x,y}^{z}>0$, and $\theta_{x,y}^{z}=0$ if $a_{x,y}^{z}\leq0$.
The spatial weighting $\alpha_{x,y}$ and channel weighting $\beta^{z}$ are defined as follows:
\begin{equation}
\begin{split}
&\alpha_{x,y}=\frac{\sum_{z=1}^{L}a_{x,y}^{z}}{(\sum_{y=1}^{H}\sum_{x=1}^{W}(\sum_{z=1}^{L}a_{x,y}^{z})^2)^\frac{1}{2}},\\
&\beta^{z}=\left\{\begin{array}{ll}
\log(\sum_{z=1}^{L}\phi^{z}/\phi^{z})&\phi^{z}>0\\
0 & \phi^{z}=0.
\end{array}\right.\end{split}
\end{equation}
thereby the final feature $\bm{f}=[f_{1},\cdots,f_{z},\cdots,f_{L}]\in\mathbb{R}^{L}$ is obtained by weight-summing:
\begin{equation}
f_{z}=\sum_{y=1}^{H}\sum_{x=1}^{W}\alpha_{x,y}\beta^{z}a_{x,y}^{z}.
\end{equation}
\end{itemize}

\subsubsection{Encoding Methods}
We employ three traditional encoding methods to aggregate feature maps into compact feature vectors: bag-of-words (BoW) \cite{Mid-BoW}, improved Fisher kernel (IFK) \cite{Hig-FV} and vector locally aggregated descriptors (VLAD) \cite{Mid-VLAD}. When processing feature maps with encoding methods, conv. activations are treated as $W\times H$ feature vectors: $\{v_{1,1},v_{1,2},\cdots,v_{x,y},\cdots,v_{W,H}\}$, here $v_{x,y}=[a_{x,y}^{1},\cdots,a_{x,y}^{L}]\in\mathbb{R}^{L}$.

\begin{itemize}
\item[-]\emph{BoW \cite{Mid-BoW}}:
BoW describes image information with statistics on the spatial distribution of local feature vectors. A codebook of $k$ centroids $\{c_{1},\cdots,c_{i},\cdots,c_{k}\}\in\mathbb{R}^{L\times k}$, for $i=1,...,k$, is learned from local feature set $\{v_{1,1},v_{1,2},\cdots,v_{W,H}\}$ via k-means clustering, then every local feature is assigned to its closest centroid. The output of BoW is a $k$-dimensional vector $\bm{f}=[f_{1},\cdots,f_{i},\cdots,f_{k}]\in\mathbb{R}^{K}$, where $f_{i}$ denotes the amount of local features that is assigned to $c_{i}$.
\item[-]\emph{IFK \cite{Hig-FV}}:
IFK is a combination of generative and discriminative approaches, it utilizes Gaussian mixture model (GMM) with $k$ Gaussian components to construct a probability density distribution of local features. Parameters of GMM is denoted as $\lambda=\{\omega_{i},\mu_{i},\Sigma_{i}\},i=1,...,k$, where $\omega_{i}$, $\mu_{i}$ and $\Sigma_{i}$ are respectively the mixture weight, mean vector and covariance matrix of Gaussian distributions. Then the $L$-dimensional gradient vectors $\mathcal{G}_{\mu,i}$ and $\mathcal{G}_{\Sigma,i}$, which are separately with respect to the mean vector and covariance matrix of $i$-th Gaussian component, are derived based on the feature set $\{v_{1,1},v_{1,2},\cdots,v_{W,H}\}$. The final IFK representation is a $2L\times k$-dimensional vector indicated as $\bm{f}=[\mathcal{G}_{\mu,1},\mathcal{G}_{\Sigma,1},\cdots,\mathcal{G}_{\mu,k},\mathcal{G}_{\Sigma,k}]\in\mathbb{R}^{2L\times k}$.
\item[-]\emph{VLAD \cite{Mid-VLAD}}:
VLAD is similar to BoW, yet it considers the statistical distribution of local features as well as the vector difference between local features and centroids simultaneously. The set of feature vectors $\{v_{1,1},v_{1,2},\cdots,v_{x,y},\cdots,v_{W,H}\}$ is clustered into a codebook $\{c_{1},\cdots,c_{i},\cdots,c_{k}\}\in\mathbb{R}^{L\times k}$ of $k$ visual words with k-means, where $i=1,...,k$. A local feature $v_{x,y}$ is assigned to its nearest visual word $c_{i}=NN(v_{x,y})$ and the vector difference $v_{x,y}-c_{i}$ between them is recorded and accumulated, ultimately, a VLAD descriptor with a dimension of $L\times k$ is represented as $\bm{f}=[\sum_{NN(v_{x,y})=c_{1}}(v_{x,y}-c_{1}),\cdots,\sum_{NN(v_{x,y})=c_{k}}(v_{x,y}-c_{k})]\in\mathbb{R}^{L\times k}$.
\end{itemize}

For both the query and reference images, before aggregating their conv. activations into global descriptors, we preprocess the feature maps with $l_{2}$-normalization. Since the dimensionality of the pooling and encoding features are different, for fair comparison, we compress aggregated feature vectors to unified dimensions with PCA dimensionality reduction. The final image representations are $l_{2}$-normalized again with the purpose of stronger robustness against noise.

\subsubsection{Multi-scale Concatenation}
We propose to utilize the fusion of conv. features derived from different scales to enhance the discriminating ability of image descriptors.

The query image and database images are firstly resized to a sequence of different scales and then separately input to CNN model to obtain multi-scale deep features. For every image, each set of feature maps are encoded into a feature vector. Finally, we simply concatenate the multi-scale feature vectors of a same original image into a long feature vector. These fused features are with multiplied dimensions due to vector concatenation, thus we compress them into a unified low dimension using PCA before retrieval.

\subsection{Scheme (II): Employing FC Features}
\subsubsection{Full-connected Features}
After removing the softmax layer, the rest portion of a CNN can be regarded as a feature extractor. In contrast with conv. layers, which can process images with any size and aspect ratio, FC layers can only process images with a fixed size. Inputting an image $\bm{I}$, FC layer straightforwardly generates single vector with a settled dimension (shown in Fig. \ref{figure:FlowChart1}), which depends on the inherent structure of CNN model.

We make use of the first two FC layers of all examined CNN models for layer comparison. To achieve higher effectiveness, we also preprocess FC activations with $l_{2}$-normalization, PCA reduction and another $l_{2}$-normalization procedure prior to retrieval.

\subsubsection{Multi-patch Pooling}
We implement multi-patch pooling method to promote the discriminating ability of deep features with multi-position information.

We crop patches with the required size of CNN models at the center and four corners from an input image. Then the horizontal, the vertical, and the horizontal-vertical reflections of these five patches are gathered, thereby, total 20 sub-patches are generated from each image. We pass those sub-patches one by one through FC layer to extract multi-patch feature vectors, which are with similar form of feature maps but without spatial distribution relationship. We aggregate such 20 feature vectors into holistic feature using pooling method, but note that SPoC and CroW can¡¯t be adopted because they are both spatial weighting based. Likewise, multi-patch pooling features are PCA reduced at end.

\subsection{Scheme (III): Fine-tuning Off-the-shelf CNN Models}
Fine-tuning is a supervised retraining process that improves the performance of CNNs for domain specific recognition tasks. The process of fine-tuning is firstly initializing the CNN model except for the softmax layer with parameters learned on the source training set, next updating the parameters of part or full CNN model via stochastic gradient descent (SGD) using the target retraining set. The dimension of the softmax layer of fine-tuned CNNs is same as the number of retraining dataset's categories, and the remaining architecture is identical with that of the original CNNs. For saving computing resources, we only fine-tune the last conv. layer of each CNN model. Specifically, for GoogLeNet, we fine-tune all 6 conv. layers between the penultimate Inception and the last Inception, as well as the last FC layer.

Concretely, we firstly change the original 1000-dimensional softmax layer into a Gaussian distributively initialized softmax layer that contains $t$ nodes, where $t$ is the number of retraining dataset's classes. Then, we perform SGD to update the weights of CNN models. The fine-tuning hyper-parameters are set as follows: epoch number 20; mini-batch size 50; momentum 0.9; initial learning rate 0.1, which is decreased to 0.05, 0.005 and 0.001 when every 5 epochs are iterated. The variation of learning rate mitigates the validation error when it tends to convergence. Moreover, dropout is applied in FC layers, and the activations are randomly set to be zero with probability 0.5 to address the problem of overfitting.

\section{Experimental Setup}
\label{sec:setup}
We perform experiments on three publicly available HRRS image datasets: RS19 \cite{Vis-2010.structural}, RSSCN7 \cite{RSSCN7} and UCM \cite{UCM}. Two standard retrieval measures are used to evaluate the results: ANMRR \cite{ANMRR} and MAP \cite{CBIR3}.

\subsection{HRRS Image Datasets}
\label{HRRSdata}
\begin{itemize}
\item[-] {RS19 \cite{Vis-2010.structural}}:The High-resolution Satellite Scene dataset is constituted by 19 categories of satellite scene images with a size of $600\times600$ pixels collected on diverse orientations and scales from Google Earth. Each category contains slightly different numbers of images for a total of 1005.
\item[-] {RSSCN7 \cite{RSSCN7}}: The Remote Sensing Scene Classification dataset is composed of 7 categories of typical scene images with a size of $400\times400$ pixels gathered from Google Earth. Each category contains 400 images, which are sampled on 4 different scales with 100 images per scale.
\item[-] {UCM \cite{UCM}}: The UC Merced Land Use/Land Cover dataset comprises 21 categories of land-use aerial images with a size of $256\times256$ pixels selected from aerial orthoimagery. Each category includes 100 images, each of which has a pixel resolution of 30cm.
\end{itemize}

In view of the relatively large scale of RSSCN7 \cite{RSSCN7}, we choose it as retraining dataset. Note that fine-tuned CNN features will be correlated with the class information of the retraining data. If such features are adopted for retrieval, the retrieval system would incline to return images with the same class label as that of the query image. However, in unsupervised retrieval task, category labels are with no practical meaning and only used for accuracy assessment. To avoid evaluation bias, we don't conduct retrieval experiment on RSSCN7 with the fine-tuned CNNs.

\subsection{Standard Retrieval Measures}
\label{ANMRR}

\begin{itemize}
\item[-] {ANMRR \cite{ANMRR}}: The average normalized modified retrieval rank (ANMRR) takes into account the number of ground truth items and the ranks obtained from the retrieval. Note that ANMRR takes values between 0 and 1, and lower value of ANMRR indicates better retrieval performance.
\item[-] {MAP \cite{CBIR3}}: The mean average precision (MAP) is the most common tool to evaluate the rank positions of all ground truth. The average precision (AvePr) for a single query image $\bm{I}$ is the mean over the precision scores of each relevant item. Different from ANMRR, the value of MAP and the performance of retrieval system are positive correlated.
\end{itemize}

\subsection{Preprocessing and Parameter Settings}
\label{parameter}
In multi-scale concatenation scheme, we define each dataset to three scales as follows: for RS19, scale1, scale2 and scale3 are $300\times300$ pixels, $600\times600$ pixels and $1200\times1200$ pixels respectively; for UCM, scale1, scale2 and scale3 are $256\times256$ pixels, $512\times512$ pixels and $1024\times1024$ pixels separately. For conv. feature aggregation, the number of K-means clustering centroids is empirically set to be 1000 and 100 respectively for BoW and VLAD, and the number of Gaussian components in the GMM for IFK is empirically set to be 100. Apart from Sec. \ref{final comparison}, the similarity measure we use in experiments is Euclidean distance.

\section{Results and Analyses}
\label{sec:results}
In this section, we present the results of experiments and analyze how the variables affect the retrieval performance. The variables include architecture of CNN model, depth of CNN layer, aggregation method for feature map, dimension of feature vector and fine-tuning.

In all of the following experiments, CNN layers are denoted using their numerical orders, such as ``\emph{conv5}'', ``\emph{conv5\_3}'' referring to conv. layers and ``\emph{fc6}'', ``\emph{fc7}'' referring to FC layers.

\begin{table*}[htp!]
\centering
\caption{Comparison of different aggregation methods. All features are compressed into 32 dimensions. The lower is the value of ANMRR the better is the accuracy and that for MAP is opposite.}
\vspace{-3mm}
\arrayrulecolor{setblue}
\settowidth\rotheadsize{\theadfont Encoding}
\renewcommand\arraystretch{1.1}
\subtable[RS19]
{
\resizebox{\textwidth}{!}{
\begin{tabular}{C{0.3cm}lC{1.1cm}C{1.1cm}cC{1.1cm}C{1.1cm}
cC{1.1cm}C{1.1cm}cC{1.1cm}C{1.1cm}cC{1.1cm}C{1.1cm}}
\toprule[1pt]
\rowcolor{setgray}
\multicolumn{2}{>{\columncolor{setgray}}c}{\textbf{Aggregation}}&
\multicolumn{2}{>{\columncolor{setgray}}c}{\textbf{CaffeNet}}   &&
\multicolumn{2}{>{\columncolor{setgray}}c}{\textbf{VGG-M}}      &&
\multicolumn{2}{>{\columncolor{setgray}}c}{\textbf{VGG-VD16}}   &&
\multicolumn{2}{>{\columncolor{setgray}}c}{\textbf{VGG-VD19}}   &&
\multicolumn{2}{>{\columncolor{setgray}}c}{\textbf{GoogLeNet}}  \\
\cmidrule[0.8pt]{3-4}\cmidrule[0.8pt]{6-7}\cmidrule[0.8pt]{9-10}
\cmidrule[0.8pt]{12-13}\cmidrule[0.8pt]{15-16}
\rowcolor{setgray}
\multicolumn{2}{>{\columncolor{setgray}}c}{\textbf{Method}}&
ANMRR&MAP(\%)&&ANMRR&MAP(\%)&&ANMRR&MAP(\%)&&
ANMRR&MAP(\%)&&ANMRR&MAP(\%)\\
\midrule[0.8pt]
\multirow{5}{*}{\rothead{Pooling}}
&\emph{Max Pooling}   &0.353&56.90&&0.316&61.49&&0.286&64.38&&0.288&64.29&&0.274&65.51\\
&\emph{Mean Pooling}  &0.389&53.03&&0.386&53.43&&0.280&65.57&&0.293&63.83&&0.250&68.42\\
&\emph{Hybrid Pooling}&0.350&57.31&&0.316&61.59&&0.284&64.65&&0.286&64.57&&0.269&66.05\\
&\emph{SPoC}          &0.411&50.53&&0.412&50.21&&0.294&64.17&&0.312&61.84&&0.263&67.18\\
&\emph{CroW }         &0.346&57.87&&0.334&59.30&&0.238&70.37&&0.246&69.39&&0.246&69.05\\
\cmidrule[0.8pt]{1-1}
\multirow{3}{*}{\rothead{Encoding}}
&\emph{BoW }          &0.319&61.22&&0.298&63.52&&0.209&73.85&&0.210&73.73&&
               \textbf{0.168}&\textbf{78.49}\\
&\emph{IFK}           &\textbf{0.244}&\textbf{69.64}&&\textbf{0.233}&\textbf{71.52}&&
      \textbf{0.190}&\textbf{76.51}&&\textbf{0.188}&\textbf{76.59}&&0.174&77.93\\
&\emph{VLAD}          &0.270&66.35&&0.260&68.02&&0.232&71.59&&0.232&71.45&&0.277&64.84\\
\bottomrule[1pt]
\end{tabular}}
}
\subtable[RSSCN7]
{
\resizebox{\textwidth}{!}{
\begin{tabular}{C{0.3cm}lC{1.1cm}C{1.1cm}cC{1.1cm}C{1.1cm}
cC{1.1cm}C{1.1cm}cC{1.1cm}C{1.1cm}cC{1.1cm}C{1.1cm}}
\toprule[1pt]
\rowcolor{setgray}
\multicolumn{2}{>{\columncolor{setgray}}c}{\textbf{Aggregation}}&
\multicolumn{2}{>{\columncolor{setgray}}c}{\textbf{CaffeNet}}   &&
\multicolumn{2}{>{\columncolor{setgray}}c}{\textbf{VGG-M}}      &&
\multicolumn{2}{>{\columncolor{setgray}}c}{\textbf{VGG-VD16}}   &&
\multicolumn{2}{>{\columncolor{setgray}}c}{\textbf{VGG-VD19}}   &&
\multicolumn{2}{>{\columncolor{setgray}}c}{\textbf{GoogLeNet}}  \\
\cmidrule[0.8pt]{3-4}\cmidrule[0.8pt]{6-7}\cmidrule[0.8pt]{9-10}
\cmidrule[0.8pt]{12-13}\cmidrule[0.8pt]{15-16}
\rowcolor{setgray}
\multicolumn{2}{>{\columncolor{setgray}}c}{\textbf{Method}}&
ANMRR&MAP(\%)&&ANMRR&MAP(\%)&&ANMRR&MAP(\%)&&
ANMRR&MAP(\%)&&ANMRR&MAP(\%)\\
\midrule[0.8pt]
\multirow{5}{*}{\rothead{Pooling}}
&\emph{Max Pooling}   &0.422&46.27&&0.396&49.41&&0.408&47.51&&0.403&47.94&&0.388&49.94\\
&\emph{Mean Pooling}  &0.388&50.10&&0.377&51.03&&0.394&49.22&&0.382&50.44&&0.367&52.42\\
&\emph{Hybrid Pooling}&0.420&46.56&&0.396&49.44&&0.407&47.61&&0.402&48.05&&0.386&50.17\\
&\emph{SPoC }         &0.387&49.94&&0.382&50.17&&0.392&48.82&&0.383&49.71&&0.392&49.30\\
&\emph{CroW }         &0.379&51.14&&0.398&49.02&&0.380&50.67&&0.371&51.66&&0.370&52.12\\
\cmidrule[0.8pt]{1-1}
\multirow{3}{*}{\rothead{Encoding}}
&\emph{BoW }          &0.378&51.54&&0.375&51.87&&0.368&52.29&&0.360&53.22&&0.354&53.86\\
&\emph{IFK}           &\textbf{0.345}&\textbf{55.51}&&\textbf{0.338}&
               \textbf{55.79}&&\textbf{0.352}&\textbf{54.01}&&
               \textbf{0.336}&\textbf{55.61}&&\textbf{0.346}&\textbf{54.97}\\
&\emph{VLAD }         &0.381&51.53&&0.395&49.90&&0.379&51.09&&0.376&51.52&&0.423&45.97\\
\bottomrule[1pt]
\end{tabular}}
}
\subtable[UCM]
{
\resizebox{\textwidth}{!}{
\begin{tabular}{C{0.3cm}lC{1.1cm}C{1.1cm}cC{1.1cm}C{1.1cm}
cC{1.1cm}C{1.1cm}cC{1.1cm}C{1.1cm}cC{1.1cm}C{1.1cm}}
\toprule[1pt]
\rowcolor{setgray}
\multicolumn{2}{>{\columncolor{setgray}}c}{\textbf{Aggregation}}&
\multicolumn{2}{>{\columncolor{setgray}}c}{\textbf{CaffeNet}}   &&
\multicolumn{2}{>{\columncolor{setgray}}c}{\textbf{VGG-M}}      &&
\multicolumn{2}{>{\columncolor{setgray}}c}{\textbf{VGG-VD16}}   &&
\multicolumn{2}{>{\columncolor{setgray}}c}{\textbf{VGG-VD19}}   &&
\multicolumn{2}{>{\columncolor{setgray}}c}{\textbf{GoogLeNet}}  \\
\cmidrule[0.8pt]{3-4}\cmidrule[0.8pt]{6-7}\cmidrule[0.8pt]{9-10}
\cmidrule[0.8pt]{12-13}\cmidrule[0.8pt]{15-16}
\rowcolor{setgray}
\multicolumn{2}{>{\columncolor{setgray}}c}{\textbf{Method}}&
ANMRR&MAP(\%)&&ANMRR&MAP(\%)&&ANMRR&MAP(\%)&&
ANMRR&MAP(\%)&&ANMRR&MAP(\%)\\
\midrule[0.8pt]
\multirow{5}{*}{\rothead{Pooling}}
&\emph{Max Pooling }  &0.469&44.92&&0.444&47.60&&0.385&53.71&&0.390&53.19&&0.387&53.13\\
&\emph{Mean Pooling } &0.535&38.75&&0.495&42.22&&0.413&50.81&&0.416&50.22&&0.381&53.94\\
&\emph{Hybrid Pooling}&0.468&45.05&&0.443&47.67&&0.384&53.83&&0.389&53.29&&0.384&53.49\\
&\emph{SPoC }         &0.532&38.61&&0.477&43.39&&0.384&53.25&&0.385&52.79&&
               \textbf{0.339}&\textbf{58.50}\\
&\emph{CroW }         &0.493&42.85&&0.473&44.63&&0.376&54.94&&0.379&54.36&&0.349&57.26\\
\cmidrule[0.8pt]{1-1}
\multirow{3}{*}{\rothead{Encoding}}
&\emph{BoW }          &0.485&43.52&&0.450&46.71&&0.372&55.34&&0.371&55.19&&0.349&57.44\\
&\emph{IFK  }         &\textbf{0.422}&\textbf{50.27}&&\textbf{0.417}&\textbf{50.40}&&
      \textbf{0.343}&\textbf{58.30}&&\textbf{0.351}&\textbf{57.58}&&0.367&55.03\\
&\emph{VLAD }         &0.484&43.49&&0.471&44.41&&0.425&49.35&&0.414&50.39&&0.498&39.85\\
\bottomrule[1pt]
\end{tabular}}
}
\label{table:I}
\end{table*}

\subsection{Convolutional Layers}
\subsubsection{Aggregation Method}
We examine the performance of different aggregation methods for conv. layers. All feature vectors are compressed to be 32-dimensional using PCA before similarity calculation.

The performance comparisons of different aggregation methods and different CNN models are shown in Table \ref{table:I}. It can be clearly observed that IFK outstands among all aggregation methods, and GoogLeNet normally outperforms other CNNs based on its deep architecture.

\subsubsection{Dimensionality Reduction}
We use hybrid pooling and IFK on feature maps generated from the last conv. layers, and then reduce each feature vector with PCA to some continuously changed dimensions: $\{8, 16, 32, 64, 128, 256, 512,\cdots\}$. The dimension of hybrid pooling features is the maximum among all pooling features (it is the concatenation of max pooling and mean pooling) so that hybrid pooling enables us to test on a wider range of varying dimensions.

\begin{figure*}[tp!]
\centering
\includegraphics[width=1\linewidth]
{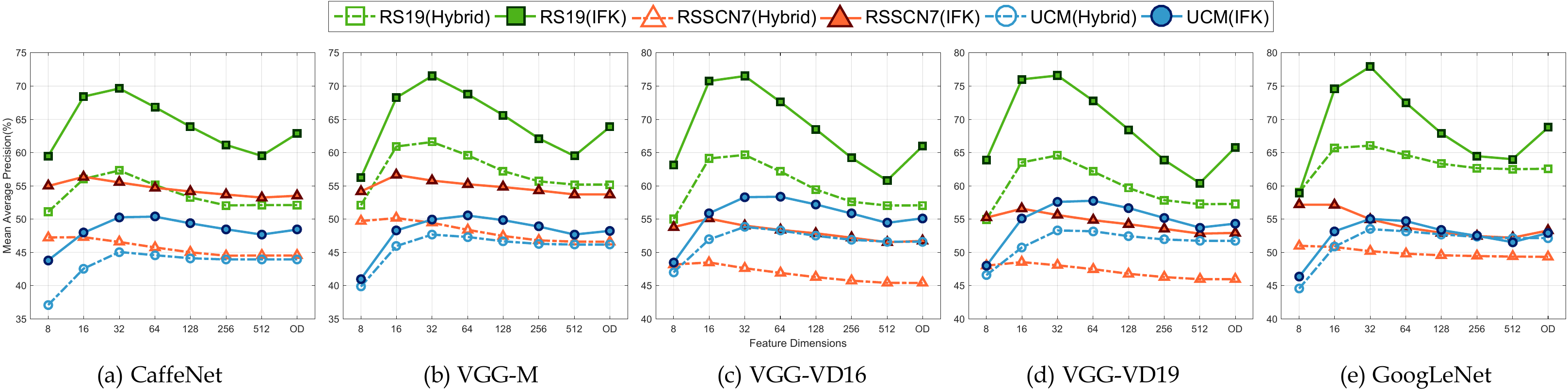}
\vspace{-4mm}
\caption{Performance of varying feature dimensions of both hybrid pooling and IFK. The best results are obtained with feature dimensions in the range of 16-64, showing that PCA compression can promote retrieval accuracy and reduce computational cost.}
\label{figure:II}
\end{figure*}

We plot the change curves of MAP for different PCA compression rates in Fig. \ref{figure:II}, where ``OD'' denotes that PCA compression is not performed. Apparently, the best accuracies of all datasets and methods are achieved in the range of 16-64 dimensions, since the redundant information is discarded along with the secondary components.

\subsubsection{Depth of Conv. Layer}
\begin{figure*}[t]
\centering
\includegraphics[width=1\linewidth]
{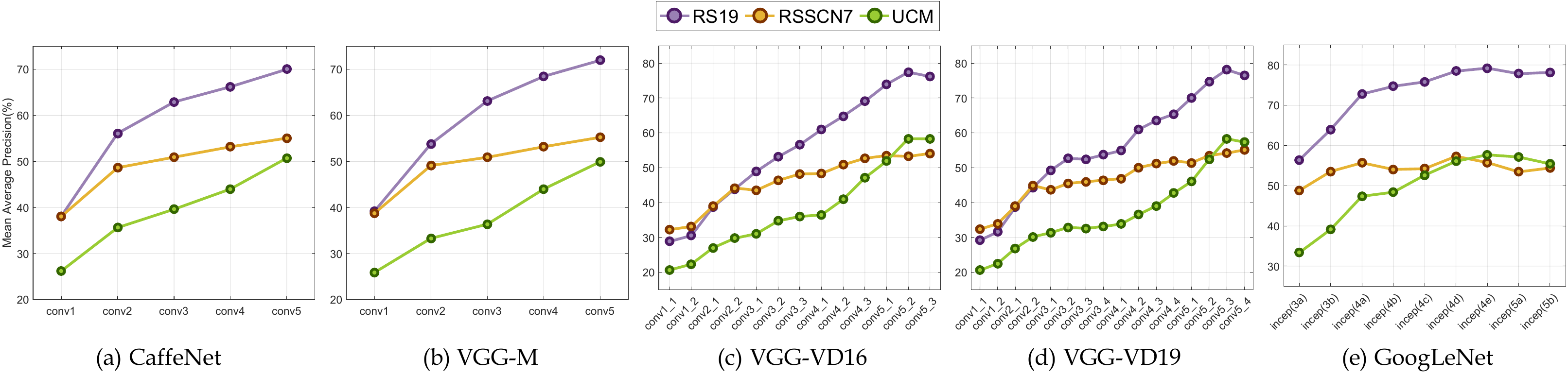}
\vspace{-4mm}
\caption{Performance of different conv. layers. Feature maps are aggregated with IFK, and the final feature vectors are compressed to 32 dimensions. It can be observed that intermediate or higher conv. layers produce better results.}
\label{figure:III}
\end{figure*}

We use IFK to encode feature maps extracted from conv. layers and compress the feature vectors to 32 dimensions uniformly. Fig. \ref{figure:III} shows MAP value of the corresponding conv. layers.

It demonstrates that deeper layers usually perform better, since activations obtained from deeper layers correspond to bigger receptive fields, which contain more information of the original image. But different from image classification, performance of retrieval is not always optimized by deeper layers, especially for the CNNs which are very deep in structure. This is because the receptive fields of extremely deep layers are with considerable large scales and unable to grasp the image details.

\subsection{Full-connected Layers}
\subsubsection{Dimensionality Reduction}
\begin{figure*}[h]
\centering
\includegraphics[width=1\linewidth]
{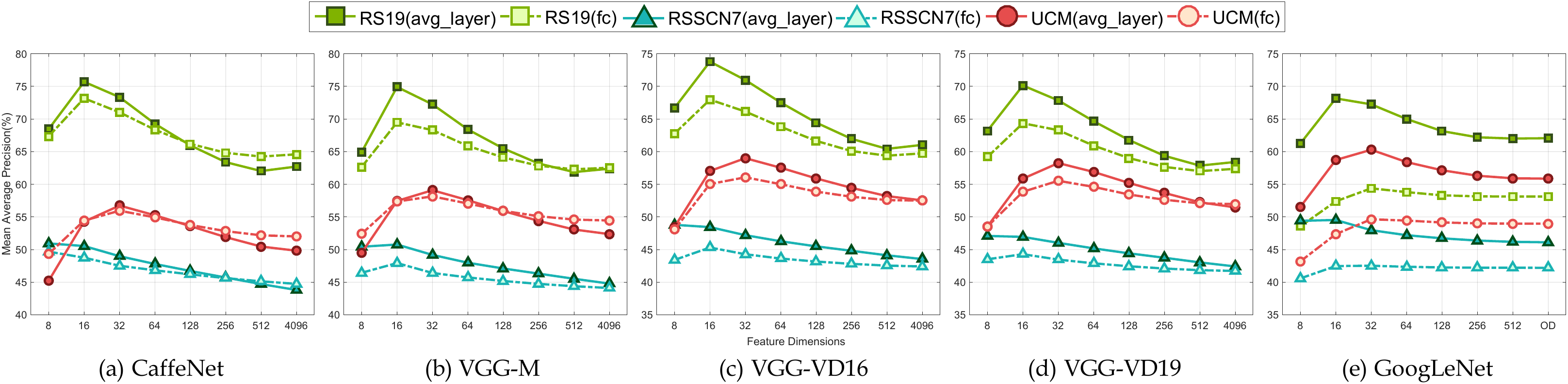}
\vspace{-4mm}
\caption{Performance of different FC layers with different dimensions. ``OD'' means the original dimensions, which are 1024 and 1000 for \emph{avg\_layer} and \emph{fc} separately. PCA reduction also helps for accuracy promotion for FC layer, and the highest MAP is achieved by the relatively lower FC layers.}
\label{figure:IV}
\end{figure*}

We as well investigate the effect of dimensionality reduction on FC features. It can be seen in Fig. \ref{figure:IV} that the optimized dimensions of all datasets are in the range of 8-32. This demonstrates that PCA compression is also effective for FC features in performance improvement.

\subsubsection{Depth of FC Layer}
As with conv. layers, we perform retrieval test on FC layers. Fig. \ref{figure:IV} clearly demonstrates that the peak of MAP of every dataset is achieved by the lower layer whatever the CNN model is. It is verified again that the deeper layers are not always the better since representations from deeper layers may be too semantically specific to the pre-training natural dataset.

\subsubsection{Convolutional vs Full-connected Layers}
We pick out layers offering the highest MAP from both conv. and FC layers, as well as the dimension which wins out for each selected layer. A comprehensive assessment is presented in Table \ref{table:II}. The aggregation method is IFK. The results show that, in most cases, features from conv. layers are more outstanding on the RS19 and RSSCN7 datasets though features from FC layers works better on UCM.

\begin{figure}[htp!]
\centering
\subfigure[RS19]
{\includegraphics[width=1\textwidth]
{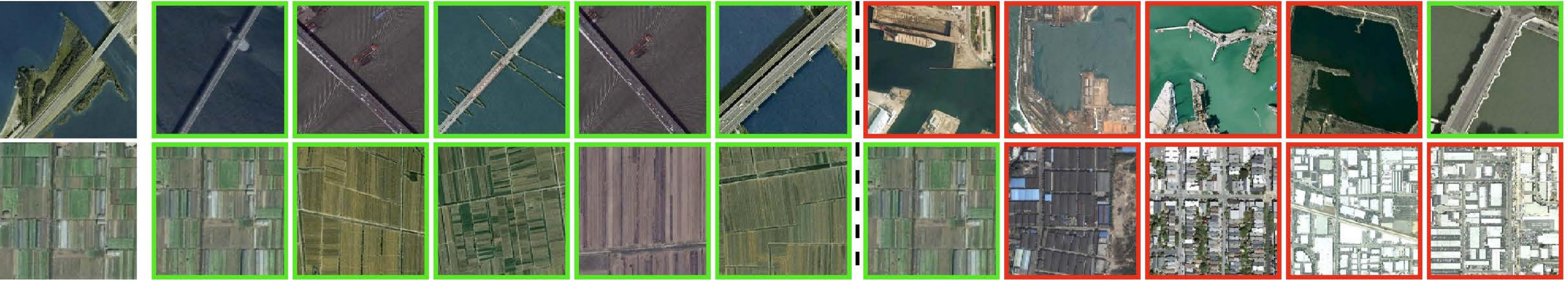}}
\vspace{-2mm}
\subfigure[RSSCN7]
{\includegraphics[width=1\textwidth]
{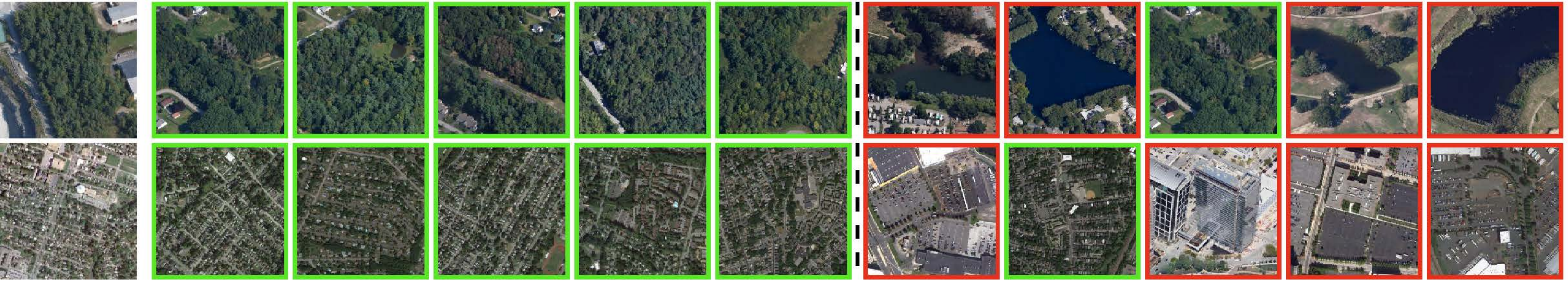}}
\vspace{-2mm}
\subfigure[UCM]
{\includegraphics[width=1\textwidth]
{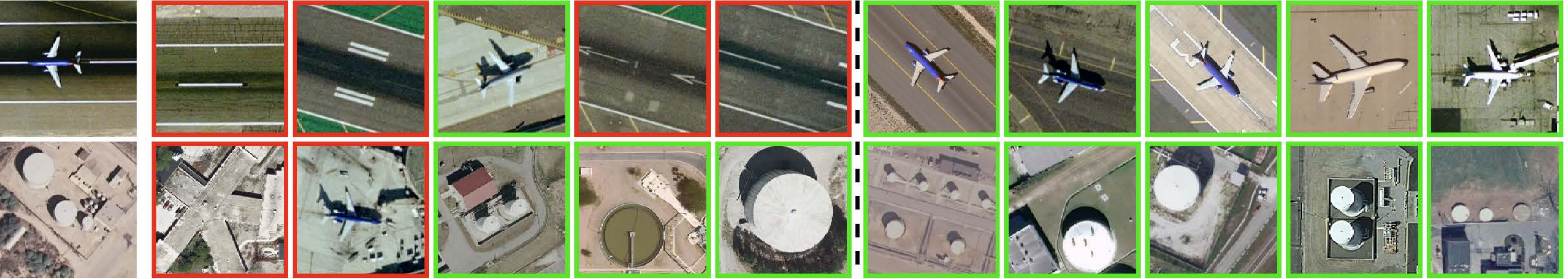}}
\vspace{-2mm}
\caption{Retrieval results produced by conv. and FC layers of GoogLeNet.}
\label{figure:VI}
\end{figure}

For further verification, we separately show top 5 images retrieved by the best conv. and FC layers of GoogLeNet at the opposite sides of a dotted line in Fig. \ref{figure:VI}. Where correct results are surrounded by green rectangle while red denotes wrong.

The results can be explained based on the characteristics of the data. Query images from RS19 and RSSCN7 are mainly covered by texture and massive structure, for example, blocky structure in farmland and canopy texture in forest. However, the most important elements in query images from UCM are ground targets: airplane and storage tanks. FC layers focus on global semantic information, whereas conv. layers extract the information from local regions. If we take away airplanes from UCM images in Fig. \ref{figure:VI}(c), the results of conv. layers can be regarded to be better than FC layers on account of the very similar runway background. Since conv. features describe the structured information better than abstract semantic information, they are inadaptable to object-oriented dataset, such as UCM.

\begin{table*}[htp!]
\caption{Comparisons between conv. and FC layers. Feature maps of conv. layers are aggregated by IFK. The lower is the value of ANMRR the better is the accuracy, that for MAP is opposite.}
\vspace{-3mm}
\arrayrulecolor{setblue}
\resizebox{\textwidth}{!}{
\begin{tabular}{lL{1.5cm}C{0.5cm}C{1.1cm}C{1.1cm}
cL{1.5cm}C{0.5cm}C{1.1cm}C{1.1cm}cL{1.5cm}C{0.5cm}C{1.1cm}C{1.1cm}}
\toprule[1.2pt]
\rowcolor{setgray}
&\multicolumn{4}{>{\columncolor{setgray}}c}{\textbf{RS19}}&
&\multicolumn{4}{>{\columncolor{setgray}}c}{\textbf{RSSCN7}}&
&\multicolumn{4}{>{\columncolor{setgray}}c}{\textbf{UCM }}\\
\cmidrule[0.6pt]{2-5}\cmidrule[0.6pt]{7-10}\cmidrule[0.6pt]{12-15}
\rowcolor{setgray}
\multirow{-2}{*}{\textbf{Nets}}
&Layer&Dim&ANMRR&MAP(\%)&&Layer&Dim&ANMRR&MAP(\%)&&Layer&Dim&ANMRR&MAP(\%)\\
\midrule[0.8pt]
\multirow{2}{*}{CaffeNet}
&conv5&32&0.241&69.90&&conv5&16&0.341&55.78&&conv5&32&0.416&50.73\\
&fc6  &16&0.190&75.70&&fc6  &8 &0.376&50.94&&fc6  &32&0.364&56.74\\
\cmidrule[0.6pt]{1-1}
\multirow{2}{*}{VGG-M}
&conv5&32&0.230&71.94&&conv5&16&0.337&55.99&&conv5&64&0.419&50.51\\
&fc6  &16&0.197&74.91&&fc6  &16&0.383&50.77&&fc6  &32&0.340&59.09\\
\cmidrule[0.6pt]{1-1}
\multirow{2}{*}{VGG-VD16}
&conv5\_2&16&0.174&78.52&&conv5\_3&16&0.340&55.30&&conv5\_2&32&0.350&58.34\\
&fc6     &16&0.203&73.79&&fc6     &8 &0.393&48.79&&fc6     &32&0.339&59.00\\
\cmidrule[0.6pt]{1-1}
\multirow{2}{*}{VGG-VD19}
&conv5\_3&16&\textbf{0.163}&\textbf{79.48}&&conv5\_4&16&0.331&56.12
&&conv5\_3&64&0.349&58.44\\
&fc6     &16&0.232&70.14&&fc6     &8 &0.409&47.09&&fc6     &32&0.345&58.25\\
\cmidrule[0.6pt]{1-1}
\multirow{2}{*}{GoogLeNet}&incep(4e)
&32&0.165&79.24&&incep(4d) &16&\textbf{0.314}&\textbf{59.04}
&&incep(4e)&64&0.349&57.97\\
&avg\_layer&16&0.243&68.16&&avg\_layer&16&0.387&49.53
&&avg\_layer&32&\textbf{0.320}&\textbf{60.29}\\
\bottomrule[1.2pt]
\end{tabular}}
\label{table:II}
\end{table*}

\subsection{Fine-tuning CNN Model}
We retrain the last conv. layer and all FC layers of CNNs using RSSCN7 and then test modified models on RS19 and UCM.

The quantitative evaluation is shown in Table \ref{table:III}, we use the last conv. layer and the first FC layer of each CNN model and apply PCA to reduce image representations to 32 dimensions. IFK is the aggregation method for conv. feature maps on account of its prominent performance shown in Table \ref{table:I}. It can be observed that all fine-tuned models produce better MAP on both test datasets whether conv. layers or FC layers are used.

\begin{figure}[htp!]
\centering
\subfigure[RS19]
{\includegraphics[width=1\textwidth]
{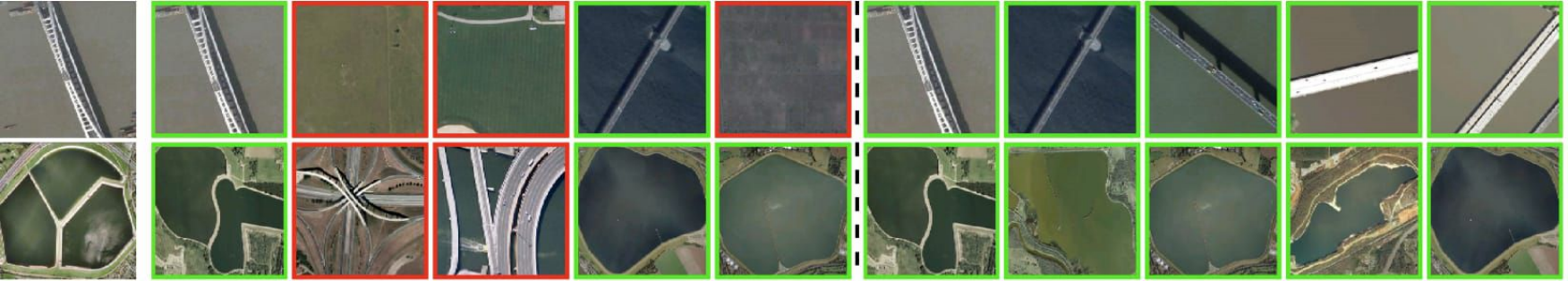}}
\vspace{-2mm}
\subfigure[UCM]
{\includegraphics[width=1\textwidth]
{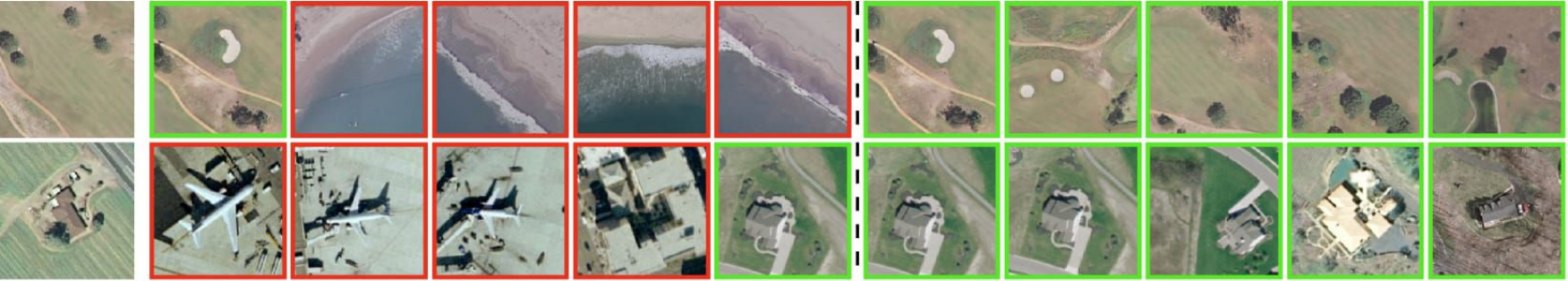}}
\vspace{-2mm}
\caption{Retrieval results produced by original and fine-tuned GoogLeNet.}
\label{figure:XI}
\end{figure}

Furthermore, we display two sets of qualitative retrieval results in Fig. \ref{figure:XI}, from left to right displayed the top 5 images retrieved with original and fine-tuned GoogLeNet. Since the best performance on RS19 and UCM is derived by different layers in Table \ref{table:III}, here we specially use \emph{inception (5b)} for RS19 and \emph{avg\_layer} for UCM. The results show that the retrained CNN model performs better.

In Fig. \ref{figure:XI}(a), it can be seen that pre-trained GoogLeNet erroneously returns a few viaduct images when querying by a pond image. This is because pond and viaduct images contain similar cross structures. And after fine-tuning GoogLeNet with HRRS images, the modified convolution filters are capable of capturing more specific semantic information from HRRS images. In Fig. \ref{figure:XI}(b), the retrieved beach images have similar curve textures with the query meadow image. Although structure and texture information can always represent the intrinsic properties of natural images, HRRS images formed by similar structure and texture are likely to contain completely different semantic properties, so that fine-tuning can reinforce the transferability of CNNs for RS image retrieval.

\begin{table*}[htp!]
\caption{Comparisons between original and fine-tuned CNNs on RS19 and UCM with MAP. IFK is applied for feature aggregating. Dimensions of all features are compressed to 32.}
\vspace{-2mm}
\arrayrulecolor{setblue}
\renewcommand\arraystretch{1.1}
\resizebox{\textwidth}{!}{
\begin{tabular}{clccccccccccc}
\toprule[1pt]
\rowcolor{setgray}
&\textbf{Layer}
&\multicolumn{2}{>{\columncolor{setgray}}c}{\textbf{CaffeNet}}&
&\multicolumn{2}{>{\columncolor{setgray}}c}{\textbf{VGG-M}}&
&\multicolumn{2}{>{\columncolor{setgray}}c}{\textbf{VGG-VD16}}&
&\multicolumn{2}{>{\columncolor{setgray}}c}{\textbf{GoogLeNet}}\\
\cmidrule[0.8pt]{3-4}\cmidrule[0.8pt]{6-7}
\cmidrule[0.8pt]{9-10}\cmidrule[0.8pt]{12-13}
\rowcolor{setgray}
\multirow{-2}{*}{\textbf{Dataset}}&\textbf{Type}
&Original&Finetuned&&Original&Finetuned&
&Original&Finetuned&&Original&Finetuned\\
\midrule[0.8pt]
\multirow{2}{*}{RS19}
&Conv.&69.90&70.13&&71.94&72.51&&76.18&77.20&&78.16&\textbf{80.21}\\
&FC &73.35&75.47&&72.25&75.55&&70.97&75.79&&67.25&72.96\\
\cmidrule[0.8pt]{1-1}
\multirow{2}{*}{UCM}
&Conv.&50.73&51.71&&49.91&52.06&&58.30&59.96&&55.44&57.82\\
&FC &56.74&58.93&&59.09&61.99&&59.00&61.97&&60.29&\textbf{62.23}\\
\bottomrule[1pt]
\end{tabular}}
\label{table:III}
\end{table*}


\subsection{Multi-scale Concatenation and Multi-patch Pooling}
\subsubsection{Multi-scale Concatenation}
We resize both query image and reference images into continuously varying scales: scale1, scale2, and scale3, and then pass them through fine-tuned GoogLeNet to obtain multi-scale feature maps from \emph{inception(5b)}. And features with different scales are concatenated for more informative image representations, referred to as scale(1,2), scale(1,3), scale(2,3) and scale(1,2,3). Take note that the encoding method we apply here is BoW instead of IFK. This is because BOW achieves great performance similar to IFK for GoogLeNet in Table \ref{table:I} and generates feature vectors with a much lower dimension (for \emph{inception(5b)}, IFK features have a dimension of 204800, whereas BoW features are 1000-dimensional). Single and multiple scale feature vectors are uniformly compressed to be 32-dimensional by PCA.

Table \ref{table:IV} presents resulting MAP on RS19 and UCM. An explanation for the experimental phenomena is that the scale of receptive field varies with the scale of input images. CNN activations focus more on global information of images with finer scale, but capturing local details from large scale. It makes sense again that RS19 and UCM are quite different in characteristics. RS19 is sensitive to low level visual features, such as edges, texture and graph structure, while UCM tends to be object-oriented. Hence the combination with enlarged scales detracts the discriminating ability of image representations on UCM.

\begin{table*}[htp!]
\caption{Performance of multi-scale concatenation method applied on \emph{inception (5b)} of fine-tuned GoogleNet.}
\vspace{-2mm}
\arrayrulecolor{setblue}
\renewcommand\arraystretch{1.1}
\begin{tabular}{ccccccccc}
\toprule[1pt]
\rowcolor{setgray}
&\multicolumn{3}{>{\columncolor{setgray}}c}{\textbf{Single Scale}}&
&\multicolumn{4}{>{\columncolor{setgray}}c}{\textbf{Multiple Scales}}\\
\cmidrule[0.8pt]{2-4}\cmidrule[0.8pt]{6-9}
\rowcolor{setgray}
\multirow{-2}{*}{\textbf{Dataset}}&
Scale1&Scale2&Scale3&&Scale(1,2)&Scale(1,3)&Scale(2,3)&Scale(1,2,3)\\
\midrule[0.8pt]
RS19&78.95&81.08&68.47&&\textbf{82.53}&69.67&72.84&73.70\\
\cmidrule[0.8pt]{1-1}
UCM &\textbf{58.69}&50.04&35.30&&52.97&36.14&37.83&38.49\\
\bottomrule[1pt]
\end{tabular}
\label{table:IV}
\end{table*}

\subsubsection{Multi-patch Pooling}
We crop 20 sub-patches of $224\times224$ pixels from corner and center of each input image and extract deep features from \emph{avg\_layer} of retrained GoogLeNet. Sets of feature vectors extracted from different locations are aggregated into compact representations via max pooling, mean pooling and hybrid pooling. All features are reduced into a dimension of 32 using PCA.

As displayed in Table \ref{table:V}, multi-patch pooling significantly boosts MAP value compared to the second column, which shows retrieval accuracies of single feature vectors. And the best results on RS19 and UCM are both acquired by mean pooling.

\begin{table*}[h]
\caption{Performance of multi-patch pooling method applied on \emph{avg\_layer} of fine-tuned GoogleNet.}
\vspace{-2mm}
\arrayrulecolor{setblue}
\renewcommand\arraystretch{1.1}
\begin{tabular}{cccccc}
\toprule[1pt]
\rowcolor{setgray}
&\textbf{Full-size Image}&&
\multicolumn{3}{>{\columncolor{setgray}}c}{\textbf{Multiple Patches}}\\
\cmidrule[0.8pt]{2-2}\cmidrule[0.8pt]{4-6}
\rowcolor{setgray}
\multirow{-2}{*}{\textbf{Dataset}}&
Single feature&&\emph{Max pooling}&\emph{Mean pooling}&\emph{Hybrid pooling}\\
\midrule[0.8pt]
RS19&72.96&&75.62&\textbf{76.80}&76.01\\
\cmidrule[0.8pt]{1-1}
UCM &62.23&&63.57&\textbf{64.56}&63.93\\
\bottomrule[1pt]
\end{tabular}
\label{table:V}
\end{table*}
\subsection{Comparison with the current methods}
We compare our proposed schemes with the recent HRRS image retrieval methods in Table \ref{table:VI}. Because relative works are almost all assessed on UCM, we only compare the accuracies on UCM. We select methods yielding the highest MAP on UCM from Table \ref{table:V} and Table \ref{table:IV}. Comparative methods are based on hand-crafted features or basic deep features.

We evaluate the performance of retrieval using ANMRR on several distance metrics: Euclidean, Cosine, Manhattan and $\chi^{2}$-square. On account of $\chi^{2}$-square distance's computational condition that elements of the feature vectors must be non-negative, it cannot be used for features that are compressed by PCA.

It is can be seen that our methods outperform all of others. Especially, the overall best accuracy is acquired by compressed multi-patch mean pooling with Euclidean distance, achieving ANMRR value of 0.285, which is about 1.4\% better than the recent CNN-based method \cite{Hig-2016.learning}. Apart from the precision, the feature dimension of our method is the lowest, which notably decreases the computation cost.

\label{final comparison}
\begin{table*}[h]
\caption{Comparison with the current methods. Several distance measures are evaluated with ANMRR, which indicates better performance with lower value.}
\vspace{-2mm}
\arrayrulecolor{setblue}
\renewcommand\arraystretch{1.1}
\begin{tabular}{llcccc}
\toprule[1pt]
\rowcolor{setgray}
&&\multicolumn{4}{>{\columncolor{setgray}}c}{\textbf{Similarity Metrics}}\\
\cmidrule[0.8pt]{3-6}
\rowcolor{setgray}
\multirow{-2}{*}{\textbf{Descriptors}}&\multirow{-2}{*}{\textbf{Dim}}
&Euclidean&Cosine&Manhattan&Chi-square\\
\midrule[0.8pt]
CCH+RIT+FPS$_{1}$+FPS$_{2}$\cite{Vis-2014.morphological}
&62   &0.640&-    &0.589&0.575\\
CCH+RIT (BoW)              \cite{Mid-2014.bag}
&128  &0.640&-    &0.613&0.585\\
Salient SIFT (BoW)             \cite{Mid-2013.geographic}
&128  &0.607&0.607&0.591&0.599\\
Dense SIFT(VLAD)           \cite{Mid-2014.performance}
&25600&-    &0.460&-    &-    \\
Pyramid LPS-aug           \cite{Mid-2016.retrieval}
&-    &0.472&-    &-    &-    \\
Manual RF VGG-M            \cite{Hig-2017.visual}
&4096 &0.316&0.316&0.333&0.315\\
Fine-tuned VGG-M           \cite{Hig-2016.learning}
&4096 &0.299&-    &-    &-\\
\cmidrule[0.8pt]{1-1}
GoogLeNet(finetuned)+BoW	    &1000 &0.423&0.423&0.685&0.639\\
GoogLeNet(finetuned)+MultiPatch	&1024 &0.314&0.314&0.323&\textbf{0.309}\\
GoogLeNet(finetuned)+BoW+PCA	&32	 &0.335&0.335&0.337&-    \\
GoogLeNet(finetuned)+MultiPatch+PCA
&32   &\textbf{0.285}&\textbf{0.285}&\textbf{0.303}&-    \\
\bottomrule[1pt]
\end{tabular}
\label{table:VI}
\end{table*}
\section{Conclusion}
\label{sec:conclusion}
We comprehensively reviewed the existing research works on content-based RS image retrieval and explored how to use CNNs to address this issue with systematical experiments. We took exhaustive influencing variables into account and performed experiments on three public HRRS image datasets with five representative CNN models. By optimizing and analyzing these variables, we achieved outstanding retrieval performance on the examined HRRS image datasets and drawn many instructive conclusions.

\bibliographystyle{IEEEtran}
\bibliography{reference}

\begin{thebibliography}{100}
\providecommand{\url}[1]{#1}
\csname url@samestyle\endcsname
\providecommand{\newblock}{\relax}
\providecommand{\bibinfo}[2]{#2}
\providecommand{\BIBentrySTDinterwordspacing}{\spaceskip=0pt\relax}
\providecommand{\BIBentryALTinterwordstretchfactor}{4}
\providecommand{\BIBentryALTinterwordspacing}{\spaceskip=\fontdimen2\font plus
\BIBentryALTinterwordstretchfactor\fontdimen3\font minus
  \fontdimen4\font\relax}
\providecommand{\BIBforeignlanguage}[2]{{%
\expandafter\ifx\csname l@#1\endcsname\relax
\typeout{** WARNING: IEEEtran.bst: No hyphenation pattern has been}%
\typeout{** loaded for the language `#1'. Using the pattern for}%
\typeout{** the default language instead.}%
\else
\language=\csname l@#1\endcsname
\fi
#2}}
\providecommand{\BIBdecl}{\relax}
\BIBdecl

\bibitem{Int-2002.knowledge-driven}
M.~Datcu, K.~Seidel, S.~D'Elia, and P.~Marchetti, ``Knowledge-driven
  information mining in remote-sensing image archives.'' \emph{E. S. A.
  Bulletin}, no. 110, pp. 26--33, 2002.

\bibitem{Int-2003.information.I}
M.~Datcu, H.~Daschiel, A.~Pelizzari, M.~Quartulli, A.~Galoppo,
  A.~Colapicchioni, M.~Pastori, K.~Seidel, P.~G. Marchetti, and S.~d'Elia,
  ``Information mining in remote sensing image archives: system concepts,''
  \emph{{IEEE} Trans. Geoscience and Remote Sensing}, vol.~41, no.~12, pp.
  2923--2936, 2003.

\bibitem{Int-2005.information.II}
H.~Daschiel and M.~Datcu, ``Information mining in remote sensing image
  archives: system evaluation,'' \emph{{IEEE} Trans. Geoscience and Remote
  Sensing}, vol.~43, no.~1, pp. 188--199, 2005.

\bibitem{Int-2013.a.review}
M.~Quartulli and I.~G. Olaizola, ``A review of eo image information mining,''
  \emph{ISPRS Journal of Photogrammetry and Remote Sensing}, vol.~75, pp.
  11--28, 2013.

\bibitem{Int-2007.Introduction}
M.~Datcu, S.~d'Elia, R.~L. King, and L.~Bruzzone, ``Introduction to the special
  section on image information mining for earth observation data,'' \emph{IEEE
  Transactions on Geoscience and Remote Sensing}, vol.~45, no.~4, pp. 795--798,
  2007.

\bibitem{Int-2010.Introduction}
M.~Datcu, R.~L. King, and S.~D'Elia, ``Introduction to the special issue on
  image information mining: Pursuing automation of geospatial intelligence for
  environment and security,'' \emph{IEEE Geoscience and Remote Sensing
  Letters}, vol.~7, no.~1, pp. 3--6, 2010.

\bibitem{Int-2017.PatternNet}
W.~Zhou, S.~Newsam, C.~Li, and Z.~Shao, ``Patternnet: A benchmark dataset for
  performance evaluation of remote sensing image retrieval,'' \emph{ISPRS
  Journal of Photogrammetry and Remote Sensing}, vol. 145, pp. 197--209, 2018.

\bibitem{Int-1993.the.intelligent}
H.~Lotz-Iwen and W.~Steinborn, ``The intelligent satellite-image information
  system isis,'' in \emph{AIP Conference Proceedings}, vol. 283, no.~1, 1993,
  pp. 727--734.

\bibitem{Int-1997.titan}
C.~Chang, B.~Moon, A.~Acharya, C.~Shock, A.~Sussman, and J.~H. Saltz, ``Titan:
  {A} high-performance remote sensing database,'' in \emph{Proceedings of the
  Thirteenth International Conference on Data Engineering}, 1997, pp. 375--384.

\bibitem{Int-1998.GeoBrowse}
G.~B. Marchisio, W.-H. Li, M.~Sannella, and J.~R. Goldschneider, ``Geobrowse:
  an integrated environment for satellite image retrieval and mining,'' in
  \emph{IGARSS'98}.

\bibitem{Int-1994.efficient}
C.~Faloutsos, R.~Barber, M.~Flickner, J.~Hafner, W.~Niblack, D.~Petkovic, and
  W.~Equitz, ``Efficient and effective querying by image content,'' \emph{J.
  Intell. Inf. Syst.}, vol.~3, no. 3/4, pp. 231--262, 1994.

\bibitem{Int-1995.content-based.image}
V.~N. Gudivada and V.~V. Raghavan, ``Content-based image retrieval systems -
  guest editors' introduction,'' \emph{{IEEE} Computer}, vol.~28, no.~9, pp.
  18--22, 1995.

\bibitem{Int-1997.new}
K.~Seidel, R.~Mastropietro, and M.~Datcu, ``New architectures for remote
  sensing image archives,'' in \emph{IGARSS'97}.

\bibitem{Int-1998.query}
K.~Seidel, M.~Schroder, H.~Rehrauer, G.~Schwarz, and M.~Datcu, ``Query by image
  content from remote sensing archives,'' in \emph{IGARSS'98}.

\bibitem{Int-2000.Image}
M.~Datcu and K.~Seidel, ``Image information mining: exploration of image
  content in large archives,'' in \emph{Aerospace Conference Proceedings},
  vol.~3.\hskip 1em plus 0.5em minus 0.4em\relax IEEE, 2000, pp. 253--264.

\bibitem{Int-2002.VisiMine}
K.~Koperski, G.~Marchisio, S.~Aksoy, and C.~Tusk, ``Visimine: Interactive
  mining in image databases,'' in \emph{IGARSS'02}.

\bibitem{Vis-2005.large-scale}
K.~W. Tobin, B.~L. Bhaduri, E.~A. Bright, A.~M. Cheriyadat, T.~P. Karnowski,
  P.~J. Palathingal, T.~E. Potok, and J.~R. Price, ``Large-scale geospatial
  indexing for image-based retrieval and analysis,'' in \emph{{ISVC} 2005}.

\bibitem{Vis-2006.automated}
K.~W. Tobin, B.~L. Bhaduri, E.~A. Bright, A.~Cheriyadat, T.~P. Karnowski, P.~J.
  Palathingal, T.~E. Potok, and J.~R. Price, ``Automated feature generation in
  large-scale geospatial libraries for content-based indexing,''
  \emph{Photogrammetric Engineering \& Remote Sensing}, vol.~72, no.~5, pp.
  531--540, 2006.

\bibitem{Vis-2006.a.framework}
M.~Klaric, G.~Scott, C.-R. Shyu, C.~Davis, and K.~Palaniappan, ``A framework
  for geospatial satellite imagery retrieval systems,'' in \emph{Proc. Int.
  Geosci. and Remote Sens. Symp}, 2006, pp. 2457--2460.

\bibitem{Vis-2007.geoiris}
C.-R. Shyu, M.~Klaric, G.~J. Scott, A.~S. Barb, C.~H. Davis, and
  K.~Palaniappan, ``Geoiris: Geospatial information retrieval and indexing
  system¡ªcontent mining, semantics modeling, and complex queries,'' \emph{IEEE
  Transactions on geoscience and remote sensing}, vol.~45, no.~4, pp. 839--852,
  2007.

\bibitem{Int-2010.System.design}
I.~M.~G. Mu{\~n}oz and M.~Datcu, ``System design considerations for image
  information mining in large archives,'' \emph{IEEE Geoscience and Remote
  Sensing Letters}, vol.~7, no.~1, pp. 13--17, 2010.

\bibitem{CBIR1}
A.~W. Smeulders, M.~Worring, S.~Santini, A.~Gupta, and R.~Jain, ``Content-based
  image retrieval at the end of the early years,'' \emph{Pattern Analysis and
  Machine Intelligence}, vol.~22, no.~12, pp. 1349--1380, 2000.

\bibitem{CBIR3}
T.~Deselaers, D.~Keysers, and H.~Ney, ``Features for image retrieval: an
  experimental comparison,'' \emph{Information Retrieval}, vol.~11, no.~2, pp.
  77--107, 2008.

\bibitem{Int-1995.content-based.retrieval}
A.~Vellaikal, C.~J. Kuo, and S.~K. Dao, ``Content-based retrieval of
  remote-sensed images using vector quantization,'' in \emph{Visual Information
  Processing IV, Orlando, FL, USA, April 17, 1995}, 1995, pp. 178--189.

\bibitem{Int-1995.system}
J.~E. Barros, J.~C. French, W.~N. Martin, and P.~M. Kelly, ``System for
  indexing multispectral satellite images for efficient content-based
  retrieval,'' in \emph{IS\&T/SPIE's Symposium on Electronic Imaging: Science
  \& Technology}.\hskip 1em plus 0.5em minus 0.4em\relax International Society
  for Optics and Photonics, 1995, pp. 228--237.

\bibitem{Int-1996.retrieving}
G.~Healey and A.~Jain, ``Retrieving multispectral satellite images using
  physics-based invariant representations,'' \emph{{IEEE} Trans. Pattern Anal.
  Mach. Intell.}, vol.~18, no.~8, pp. 842--848, 1996.

\bibitem{Int-1998.meta}
K.~Seidel, M.~Schroder, H.~Rehrauer, and M.~Datcu, ``Meta features for remote
  sensing image content indexing,'' in \emph{IGARSS'98}.

\bibitem{Int-1998.spatial.I}
M.~Datcu, K.~Seidel, and M.~Walessa, ``Spatial information retrieval from
  remote-sensing images. i. information theoretical perspective,'' \emph{IEEE
  transactions on geoscience and remote sensing}, vol.~36, no.~5, pp.
  1431--1445, 1998.

\bibitem{Int-1998.spatial.II}
M.~Schroder, H.~Rehrauer, K.~Seidel, and M.~Datcu, ``Spatial information
  retrieval from remote-sensing images. ii. gibbs-markov random fields,''
  \emph{IEEE Transactions on geoscience and remote sensing}, vol.~36, no.~5,
  pp. 1446--1455, 1998.

\bibitem{Vis-2000.Gibbs}
M.~Schr{\"o}der, M.~Walessa, H.~Rehrauer, K.~Seidel, and M.~Datcu, ``Gibbs
  random field models: a toolbox for spatial information extraction,''
  \emph{Computers \& Geosciences}, vol.~26, no.~4, pp. 423--432, 2000.

\bibitem{Vis-2002.a.retrieval.system}
T.~Bretschneider and O.~Kao, ``A retrieval system for remotely sensed
  imagery,'' in \emph{International Conference on Imaging Science, Systems, and
  Technology}, 2002.

\bibitem{Vis-2002.retrieval.of.remotely}
T.~Bretschneider, R.~Cavet, and O.~Kao, ``Retrieval of remotely sensed imagery
  using spectral information content,'' in \emph{IGARSS'02}.

\bibitem{Vis-2004.remote.sensing.imagery}
Y.~Hongyu, L.~Bicheng, and C.~Wen, ``Remote sensing imagery retrieval based-on
  gabor texture feature classification,'' in \emph{ICSP'04}.

\bibitem{Vis-2004.using.texture}
S.~Newsam, L.~Wang, S.~Bhagavathy, and B.~S. Manjunath, ``Using texture to
  analyze and manage large collections of remote sensed image and video data,''
  \emph{Applied optics}, vol.~43, no.~2, pp. 210--217, 2004.

\bibitem{Vis-2005.local.shape}
A.~Ma and I.~K. Sethi, ``Local shape association based retrieval of infrared
  satellite images,'' in \emph{IEEE International Symposium on Multimedia},
  2005.

\bibitem{Vis-2007.camparing}
S.~D. Newsam and Y.~Yang, ``Comparing global and interest point descriptors for
  similarity retrieval in remote sensed imagery,'' in \emph{{ACM-GIS} 2007}.

\bibitem{Vis-1999.An.environment}
P.~Agouris, J.~Carswell, and A.~Stefanidis, ``An environment for content-based
  image retrieval from large spatial databases,'' \emph{Journal of
  Photogrammetry and Remote Sensing}, vol.~54, no.~4, pp. 263--272, 1999.

\bibitem{Vis-2001.Query-by-shape}
F.~Dell'Acqua and P.~Gamba, ``Query-by-shape in meteorological image archives
  using the point diffusion technique,'' \emph{IEEE transactions on geoscience
  and remote sensing}, vol.~39, no.~9, pp. 1834--1843, 2001.

\bibitem{Vis-2005.Wavelet.features}
V.~P. Shah, N.~H. Younan, S.~Durba, and R.~King, ``Wavelet features for
  information mining in remote sensing archives,'' in \emph{IGARSS'05}.

\bibitem{Vis-2007.A.systematic}
V.~P. Shah, N.~H. Younan, S.~S. Durbha, and R.~L. King, ``A systematic approach
  to wavelet-decomposition-level selection for image information mining from
  geospatial data archives,'' \emph{IEEE Transactions on Geoscience and Remote
  Sensing}, vol.~45, no.~4, pp. 875--878, 2007.

\bibitem{Vis-2014.Improved.color}
Z.~Shao, W.~Zhou, L.~Zhang, and J.~Hou, ``Improved color texture descriptors
  for remote sensing image retrieval,'' \emph{journal of applied remote
  sensing}, vol.~8, no.~1, pp. 083\,584--083\,584, 2014.

\bibitem{Vis-2015.Multiscale}
S.~Bouteldja and A.~Kourgli, ``Multiscale texture features for the retrieval of
  high resolution satellite images,'' in \emph{IWSSIP 2015}.

\bibitem{Int-1999.content-based}
G.~B. Marchisio and J.~Cornelison, ``Content-based search and clustering of
  remote sensing imagery,'' in \emph{IGARSS'99}.

\bibitem{Vis-1997.deriving}
C.~Li and V.~Castelli, ``Deriving texture feature set for content-based
  retrieval of satellite image database,'' in \emph{{ICIP}'97}.

\bibitem{Vis-2000.multi-level}
K.~Koperski and G.~B. Marchisio, ``Multi-level indexing and {GIS} enhanced
  learning for satellite imageries,'' in \emph{Proceedings of the International
  Workshop on Multimedia Data Mining}, 2000.

\bibitem{Vis-2004.integrated}
J.~Li and R.~M. Narayanan, ``Integrated spectral and spatial information mining
  in remote sensing imagery,'' \emph{{IEEE} Trans. Geoscience and Remote
  Sensing}, vol.~42, no.~3, pp. 673--685, 2004.

\bibitem{Vis-2004.retrieval.using.texture}
S.~D. Newsam and C.~Kamath, ``Retrieval using texture features in
  high-resolution multispectral satellite imagery,'' in \emph{Defense and
  Security}, 2004, pp. 21--32.

\bibitem{Vis-2004.semantics-based}
Y.~Li and T.~R. Bretschneider, ``Semantics-based satellite image retrieval
  using low-level features,'' in \emph{{IGARSS} 2004}.

\bibitem{Vis-2009.retrieval}
P.~Maheswary and N.~Srivastava, ``Retrieval of remote sensing images using
  colour and texture attribute,'' \emph{International Journal of Computer
  Science and Information Security}, vol.~4, no.~8, pp. 3--15, 2009.

\bibitem{Vis-2009.searching}
A.~Samal, S.~K. Bhatia, P.~Vadlamani, and D.~Marx, ``Searching satellite
  imagery with integrated measures,'' \emph{Pattern Recognition}, vol.~42,
  no.~11, pp. 2502--2513, 2009.

\bibitem{Vis-2009.Prototype.system}
P.~Maheshwary and N.~Srivastava, ``Prototype system for retrieval of remote
  sensing images based on color moment and gray level co-occurrence matrix,''
  \emph{P. Maheshwary and N. Srivastava," Prototype System for Retrieval of
  Remote Sensing Images based on Color Moment and Gray Level Co-Occurrence
  Matrix", International Journal of Computer Science Issues, IJCSI, Volume 3,
  pp20-23, August 2009}, vol.~3, 2009.

\bibitem{Vis-2014.Remote.Sensing}
Z.~Shao, W.~Zhou, and Q.~Cheng, ``Remote sensing image retrieval with combined
  features of salient region,'' \emph{The International Archives of
  Photogrammetry, Remote Sensing and Spatial Information Sciences}, vol.~40,
  no.~6, p.~83, 2014.

\bibitem{Vis-2015.Dual-tree}
H.~Sebai, A.~Kourgli, and A.~Serir, ``Dual-tree complex wavelet transform
  applied on color descriptors for remote-sensed images retrieval,''
  \emph{Journal of Applied Remote Sensing}, vol.~9, no.~1, pp.
  095\,994--095\,994, 2015.

\bibitem{Sim-2004.comparative}
Q.~Bao and P.~Guo, ``Comparative studies on similarity measures for remote
  sensing image retrieval,'' in \emph{Proceedings of the {IEEE} International
  Conference on Systems, Man {\&} Cybernetics: The Hague}.

\bibitem{Sim-2008.A.similarity}
L.~Gueguen and M.~Datcu, ``A similarity metric for retrieval of compressed
  objects: Application for mining satellite image time series,'' \emph{IEEE
  Transactions on Knowledge and Data Engineering}, vol.~20, no.~4, pp.
  562--575, 2008.

\bibitem{Sim-2012.an.endmenber}
M.~Gra{\~{n}}a and M.~A. Veganzones, ``An endmember-based distance for content
  based hyperspectral image retrieval,'' \emph{Pattern Recognition}, vol.~45,
  no.~9, pp. 3472--3489, 2012.

\bibitem{Sim-2012.dictionary}
M.~A. Veganzones, M.~Datcu, and M.~Grana, ``Dictionary based hyperspectral
  image retrieval.'' in \emph{ICPRAM (1)}, 2012, pp. 426--432.

\bibitem{Int-2000.interactive}
M.~Schr{\"{o}}der, H.~Rehrauer, K.~Seidel, and M.~Datcu, ``Interactive learning
  and probabilistic retrieval in remote sensing image archives,'' \emph{{IEEE}
  Trans. Geoscience and Remote Sensing}, vol.~38, no.~5, pp. 2288--2298, 2000.

\bibitem{Rel-2007.interactive}
M.~Ferecatu and N.~Boujemaa, ``Interactive remote-sensing image retrieval using
  active relevance feedback,'' \emph{{IEEE} Trans. Geoscience and Remote
  Sensing}, vol.~45, no.~4, pp. 818--826, 2007.

\bibitem{Rel-2015.a.novel}
B.~Demir and L.~Bruzzone, ``A novel active learning method in relevance
  feedback for content-based remote sensing image retrieval,'' \emph{{IEEE}
  Trans. Geoscience and Remote Sensing}, vol.~53, no.~5, pp. 2323--2334, 2015.

\bibitem{Rel-2001.Fast.retrieval}
I.~E. Alber, Z.~Xiong, N.~Yeager, M.~Farber, and W.~M. Pottenger, ``Fast
  retrieval of multi-and hyperspectral images using relevance feedback,'' in
  \emph{IGARSS'01}.

\bibitem{Rel-2002.Probabilistic}
S.~Aksoy, G.~Marchisio, K.~Koperski, and C.~Tusk, ``Probabilistic retrieval
  with a visual grammar,'' in \emph{IGARSS'02}.

\bibitem{Rel-2004.Interactive}
S.~Aksoy, K.~Koperski, C.~Tusk, and G.~Marchisio, ``Interactive training of
  advanced classifiers for mining remote sensing image archives,'' in \emph{ACM
  SIGKDD}, 2004.

\bibitem{Rel-2006.Scalable}
R.~Datta, J.~Li, A.~Parulekar, and J.~Z. Wang, ``Scalable remotely sensed image
  mining using supervised learning and content-based retrieval,''
  \emph{Pennsylvania State Univ., State College, PA, USA, Tech. Rep. CSE}, pp.
  06--019, 2006.

\bibitem{Rel-2007.Learning-unlearning}
M.~Costache and M.~Datcu, ``Learning-unlearning for mining high resolution eo
  images,'' in \emph{IGARSS 2007}.

\bibitem{Rel-2007.Semantic-sensitive}
Y.~Li and T.~R. Bretschneider, ``Semantic-sensitive satellite image
  retrieval,'' \emph{IEEE Transactions on Geoscience and Remote Sensing},
  vol.~45, no.~4, pp. 853--860, 2007.

\bibitem{Rel-2010.Visual.information}
A.~S. Barb and C.-R. Shyu, ``Visual information mining and ranking using graded
  relevance assessments in satellite image databases,'' in \emph{IGARSS 2010}.

\bibitem{Rel-2010.Visual-semantic}
A.~S. Barb and C.~Shyu, ``Visual-semantic modeling in content-based geospatial
  information retrieval using associative mining techniques,'' \emph{{IEEE}
  Geosci. Remote Sensing Lett.}, vol.~7, no.~1, pp. 38--42, 2010.

\bibitem{Rel-2011.An.interactive}
L.~Gueguen, M.~Pesaresi, and P.~Soille, ``An interactive image mining tool
  handling gigapixel images,'' in \emph{IGARSS 2011}.

\bibitem{Vis-2014.morphological}
E.~Aptoula, ``Remote sensing image retrieval with global morphological texture
  descriptors,'' \emph{{IEEE} Trans. Geoscience and Remote Sensing}, vol.~52,
  no.~5, pp. 3023--3034, 2014.

\bibitem{Vis-2011.entropy-balanced}
G.~J. Scott, M.~N. Klaric, C.~H. Davis, and C.~Shyu, ``Entropy-balanced bitmap
  tree for shape-based object retrieval from large-scale satellite imagery
  databases,'' \emph{{IEEE} Trans. Geoscience and Remote Sensing}, vol.~49,
  no.~5, pp. 1603--1616, 2011.

\bibitem{Mid-BoW}
J.~Sivic and A.~Zisserman, ``Video google: {A} text retrieval approach to
  object matching in videos,'' in \emph{{(ICCV} 2003)}.

\bibitem{Hig-FV}
F.~Perronnin and C.~R. Dance, ``Fisher kernels on visual vocabularies for image
  categorization,'' in \emph{{(CVPR} 2007)}.

\bibitem{Mid-VLAD}
H.~J{\'{e}}gou, M.~Douze, C.~Schmid, and P.~P{\'{e}}rez, ``Aggregating local
  descriptors into a compact image representation,'' in \emph{{CVPR} 2010}.

\bibitem{Hig-2016.learning}
W.~Zhou, S.~Newsam, C.~Li, and Z.~Shao, ``Learning low dimensional
  convolutional neural networks for high-resolution remote sensing image
  retrieval,'' \emph{Remote Sensing}, vol.~9, no.~5, p. 489, 2017.

\bibitem{Hig-2017.visual}
P.~Napoletano, ``Visual descriptors for content-based retrieval of remote
  sensing images,'' \emph{International journal of remote sensing}, vol.~39,
  no.~5, pp. 1343--1376, 2018.

\bibitem{Hig-2017.Retrieving.Aerial}
T.-B. Jiang, G.-S. Xia, Q.-K. Lu, and W.-M. Shen, ``Retrieving aerial scene
  images with learned deep image-sketch features,'' \emph{Journal of Computer
  Science and Technology}, vol.~32, no.~4, pp. 726--737, 2017.

\bibitem{AlexNet}
A.~Krizhevsky, I.~Sutskever, and G.~E. Hinton, ``Imagenet classification with
  deep convolutional neural networks,'' in \emph{Advances in neural information
  processing systems}, 2012, pp. 1097--1105.

\bibitem{appSum1}
A.~Razavian, H.~Azizpour, J.~Sullivan, and S.~Carlsson, ``Cnn features
  off-the-shelf: an astounding baseline for recognition,'' in \emph{Proceedings
  of the IEEE Conference on Computer Vision and Pattern Recognition Workshops},
  2014, pp. 806--813.

\bibitem{appSum4}
P.~Sermanet, D.~Eigen, X.~Zhang, M.~Mathieu, R.~Fergus, and Y.~LeCun,
  ``Overfeat: Integrated recognition, localization and detection using
  convolutional networks,'' in \emph{International Conference on Learning
  Representations}, 2014.

\bibitem{CBIR2}
R.~Datta, D.~Joshi, J.~Li, and J.~Z. Wang, ``Image retrieval: Ideas,
  influences, and trends of the new age,'' \emph{ACM Computing Surveys (CSUR)},
  vol.~40, no.~2, p.~5, 2008.

\bibitem{Rel-1996.Bayesian}
I.~J. Cox, M.~L. Miller, S.~M. Omohundro, and P.~N. Yianilos, ``Pichunter:
  Bayesian relevance feedback for image retrieval,'' in \emph{{ICPR} 1996}.

\bibitem{Rel-1998.mindreader}
Y.~Ishikawa, R.~Subramanya, and C.~Faloutsos, ``Mindreader: Querying databases
  through multiple examples,'' in \emph{VLDB'98}.

\bibitem{Oth-2005.Study.on}
D.~Peijun, C.~Yunhao, T.~Hong, and F.~Tao, ``Study on content-based remote
  sensing image retrieval,'' in \emph{IGARSS 2005}.

\bibitem{Vis-GLCM}
R.~M. Haralick, K.~S. Shanmugam, and I.~Dinstein, ``Textural features for image
  classification,'' \emph{{IEEE} Trans. Systems, Man, and Cybernetics}, vol.~3,
  no.~6, pp. 610--621, 1973.

\bibitem{Vis-wavelet}
S.~Mallat, ``A theory for multiresolution signal decomposition: The wavelet
  representation,'' \emph{{IEEE} Trans. Pattern Anal. Mach. Intell.}, vol.~11,
  no.~7, pp. 674--693, 1989.

\bibitem{Vis-wavelet.based.histogram}
A.~Boggess, F.~J. Narcowich, D.~L. Donoho, and P.~L. Donoho, ``A first course
  in wavelets with fourier analysis,'' \emph{Physics Today}, vol.~55, no.~5,
  p.~63, 2002.

\bibitem{Vis-Gabor}
J.~G. Daugman, ``Complete discrete 2-d gabor transforms by neural networks for
  image analysis and compression,'' \emph{{IEEE} Trans. Acoustics, Speech, and
  Signal Processing}, vol.~36, no.~7, pp. 1169--1179, 1988.

\bibitem{Vis-Gabor.coefficient}
B.~S. Manjunath and W.~Ma, ``Texture features for browsing and retrieval of
  image data,'' \emph{{IEEE} Trans. Pattern Anal. Mach. Intell.}, vol.~18,
  no.~8, pp. 837--842, 1996.

\bibitem{Vis-LBP}
M.~Pietik{\"{a}}inen, T.~Ojala, and Z.~Xu, ``Rotation-invariant texture
  classification using feature distributions,'' \emph{Pattern Recognition},
  vol.~33, no.~1, pp. 43--52, 2000.

\bibitem{Vis-SIFT}
D.~G. Lowe, ``Distinctive image features from scale-invariant keypoints,''
  \emph{International Journal of Computer Vision}, vol.~60, no.~2, pp. 91--110,
  2004.

\bibitem{Vis-topographic}
V.~Caselles, B.~Coll, and J.~Morel, ``Topographic maps and local contrast
  changes in natural images,'' \emph{International Journal of Computer Vision},
  vol.~33, no.~1, pp. 5--27, 1999.

\bibitem{Vis-indexing}
G.-S. Xia, J.~Delon, and Y.~Gousseau, ``Shape-based invariant texture
  indexing,'' \emph{International Journal of Computer Vision}, vol.~88, no.~3,
  pp. 382--403, 2010.

\bibitem{Vis-2010.structural}
G.-S. Xia, W.~Yang, J.~Delon, Y.~Gousseau, H.~Sun, and H.~Ma{\^\i}tre,
  ``Structural high-resolution satellite image indexing,'' in \emph{ISPRS TC
  VII Symposium-100 Years}, vol.~38, 2010, pp. 298--303.

\bibitem{xia2014texture}
G.~Liu, G.-S. Xia, W.~Yang, and L.~Zhang, ``Texture analysis by using shapes
  co-occurrence patterns,'' in \emph{International Conference on Pattern
  Recognition}, 2014, pp. 1--6.

\bibitem{xia2017SCOPE}
G.~S. Xia, G.~Liu, X.~Bai, and L.~Zhang, ``Texture characterization using shape
  co-occurrence patterns,'' \emph{IEEE Transactions on Image Processing},
  vol.~PP, no.~99, pp. 1--1, 2017.

\bibitem{Vis-2013.scene.semantic.matching}
M.~Wang and T.~Song, ``Remote sensing image retrieval by scene semantic
  matching,'' \emph{{IEEE} Trans. Geoscience and Remote Sensing}, vol.~51, no.
  5-1, pp. 2874--2886, 2013.

\bibitem{Mid-2013.geographic}
Y.~Yang and S.~D. Newsam, ``Geographic image retrieval using local invariant
  features,'' \emph{{IEEE} Trans. Geoscience and Remote Sensing}, vol.~51,
  no.~2, pp. 818--832, 2013.

\bibitem{Mid-2014.bag}
E.~Aptoula, ``Bag of morphological words for content-based geographical
  retrieval,'' in \emph{{CBMI} 2014}.

\bibitem{Mid-2015.An.improved}
J.~Yang, J.~Liu, and Q.~Dai, ``An improved bag-of-words framework for remote
  sensing image retrieval in large-scale image databases,'' \emph{International
  Journal of Digital Earth}, vol.~8, no.~4, pp. 273--292, 2015.

\bibitem{Mid-pattern}
P.~Maragos, ``Pattern spectrum and multiscale shape representation,''
  \emph{{IEEE} Trans. Pattern Anal. Mach. Intell.}, vol.~11, no.~7, pp.
  701--716, 1989.

\bibitem{Mid-2016.retrieval}
P.~Bosilj, E.~Aptoula, S.~Lef{\`{e}}vre, and E.~Kijak, ``Retrieval of remote
  sensing images with pattern spectra descriptors,'' \emph{{ISPRS} Int. J.
  Geo-Information}, vol.~5, no.~12, p. 228, 2016.

\bibitem{Mid-2014.performance}
S.~{\"{O}}zkan, T.~Ates, E.~Tola, M.~Soysal, and E.~Esen, ``Performance
  analysis of state-of-the-art representation methods for geographical image
  retrieval and categorization,'' \emph{{IEEE} Geosci. Remote Sensing Lett.},
  vol.~11, no.~11, pp. 1996--2000, 2014.

\bibitem{Rel-ScSPM}
J.~Yang, K.~Yu, Y.~Gong, and T.~S. Huang, ``Linear spatial pyramid matching
  using sparse coding for image classification,'' in \emph{{(CVPR} 2009)}.

\bibitem{Rel-2016.a.three-layered}
Y.~Wang, L.~Zhang, X.~Tong, L.~Zhang, Z.~Zhang, H.~Liu, X.~Xing, and P.~T.
  Mathiopoulos, ``A three-layered graph-based learning approach for remote
  sensing image retrieval,'' \emph{{IEEE} Trans. Geoscience and Remote
  Sensing}, vol.~54, no.~10, pp. 6020--6034, 2016.

\bibitem{Mid-2015.High-resolution}
W.~Zhou, Z.~Shao, C.~Diao, and Q.~Cheng, ``High-resolution remote-sensing
  imagery retrieval using sparse features by auto-encoder,'' \emph{Remote
  Sensing Letters}, vol.~6, no.~10, pp. 775--783, 2015.

\bibitem{Mid-2016.Content-Based}
Y.~Li, Y.~Zhang, C.~Tao, and H.~Zhu, ``Content-based high-resolution remote
  sensing image retrieval via unsupervised feature learning and collaborative
  affinity metric fusion,'' \emph{Remote Sensing}, vol.~8, no.~9, p. 709, 2016.

\bibitem{fea2}
M.~D. Zeiler and R.~Fergus, ``Visualizing and understanding convolutional
  networks,'' in \emph{ECCV 2014}.\hskip 1em plus 0.5em minus 0.4em\relax
  Springer, 2014, pp. 818--833.

\bibitem{Hig-AID}
G.-S. Xia, J.~Hu, F.~Hu, B.~Shi, X.~Bai, Y.~Zhong, L.~Zhang, and X.~Lu, ``Aid:
  A benchmark data set for performance evaluation of aerial scene
  classification,'' \emph{IEEE Transactions on Geoscience and Remote Sensing},
  2017.

\bibitem{Sim-metric.learning.I}
L.~Si, R.~Jin, S.~C.~H. Hoi, and M.~R. Lyu, ``Collaborative image retrieval via
  regularized metric learning,'' \emph{Multimedia Syst.}, vol.~12, no.~1, pp.
  34--44, 2006.

\bibitem{Sim-metric.learning.II}
S.~C.~H. Hoi, W.~Liu, and S.~Chang, ``Semi-supervised distance metric learning
  for collaborative image retrieval and clustering,'' \emph{{TOMCCAP}}, vol.~6,
  no.~3, pp. 18:1--18:26, 2010.

\bibitem{Sim-metric.learning.III}
C.~Huang, S.~Zhu, and K.~Yu, ``Large scale strongly supervised ensemble metric
  learning, with applications to face verification and retrieval,'' \emph{arXiv
  preprint arXiv:1212.6094}, 2012.

\bibitem{Sim-metric.learning.IV}
J.~Lee, R.~Jin, and A.~K. Jain, ``Rank-based distance metric learning: An
  application to image retrieval,'' in \emph{{(CVPR} 2008)}.

\bibitem{Sim-2016.Region-Based}
B.~Chaudhuri, B.~Demir, L.~Bruzzone, and S.~Chaudhuri, ``Region-based retrieval
  of remote sensing images using an unsupervised graph-theoretic approach,''
  \emph{IEEE Geoscience and Remote Sensing Letters}, vol.~13, no.~7, pp.
  987--991, 2016.

\bibitem{Sim-2015.NetVLAD}
R.~Arandjelovic, P.~Gron{\'{a}}t, A.~Torii, T.~Pajdla, and J.~Sivic, ``Netvlad:
  {CNN} architecture for weakly supervised place recognition,'' in \emph{IEEE
  Conference on Computer Vision and Pattern Recognition}, 2016, pp. 5297--5307.

\bibitem{Rel-automatic.manual}
R.~A. Baeza{-}Yates and B.~A. Ribeiro{-}Neto, \emph{Modern Information
  Retrieval}.\hskip 1em plus 0.5em minus 0.4em\relax {ACM} Press /
  Addison-Wesley, 1999.

\bibitem{Rel-SimpleMKL}
A.~Rakotomamonjy, F.~R. Bach, S.~Canu, and Y.~Grandvalet, ``Simplemkl,''
  \emph{Journal of Machine Learning Research}, vol.~9, no. Nov, pp. 2491--2521,
  2008.

\bibitem{Rel-1995.optimization}
C.~Buckley and G.~Salton, ``Optimization of relevance feedback weights,'' in
  \emph{SIGIR'95}.

\bibitem{Rel-2000.the.Bayesian}
I.~J. Cox, M.~L. Miller, T.~P. Minka, T.~V. Papathomas, and P.~N. Yianilos,
  ``The bayesian image retrieval system, pichunter: theory, implementation, and
  psychophysical experiments,'' \emph{{IEEE} Trans. Image Processing}, vol.~9,
  no.~1, pp. 20--37, 2000.

\bibitem{Rel-2001.support}
S.~Tong and E.~Y. Chang, ``Support vector machine active learning for image
  retrieval,'' in \emph{{ACM} 2001}.

\bibitem{Rel-2001.a.neural}
L.~Zhang, F.~Lin, and B.~Zhang, ``A neural network based self-learning
  algorithm of image retrieval,'' \emph{Chinese Journal of Software}, vol.~12,
  no.~10, pp. 1479--1485, 2001.

\bibitem{Oth-2015.distributed}
L.~Mascolo, M.~Quartulli, P.~Guccione, G.~Nico, and I.~G. Olaizola,
  ``Distributed mining of large scale remote sensing image archives on public
  computing infrastructures,'' \emph{arXiv preprint arXiv:1501.05286}.

\bibitem{Vis-polygon.I}
N.~Ramesh and I.~Sethi, ``A model based industrial part recognition system
  using hashing,'' in \emph{Proc. 22nd Intl. Symposium on Industrial Robots,
  Intl. Robots and Vision Automation Conf}, 2014.

\bibitem{Vis-polygon.II}
I.~K. Sethi and N.~Ramesh, ``Local association based recognition of
  two-dimensional objects,'' \emph{Mach. Vis. Appl.}, vol.~5, no.~4, pp.
  265--276, 1992.

\bibitem{Oth-2014.kernel-based}
B.~Demir and L.~Bruzzo, ``Kernel-based hashing for content-based image retrval
  in large remote sensing data archive,'' in \emph{IGARSS 2014}.

\bibitem{Oth-2016.hashing-based}
B.~Demir and L.~Bruzzone, ``Hashing-based scalable remote sensing image search
  and retrieval in large archives,'' \emph{{IEEE} Trans. Geoscience and Remote
  Sensing}, vol.~54, no.~2, pp. 892--904, 2016.

\bibitem{CaffeNet}
Y.~Jia, E.~Shelhamer, J.~Donahue, S.~Karayev, J.~Long, R.~Girshick,
  S.~Guadarrama, and T.~Darrell, ``Caffe: Convolutional architecture for fast
  feature embedding,'' in \emph{Proceedings of the ACM International Conference
  on Multimedia}, 2014, pp. 675--678.

\bibitem{VGGNets}
K.~Chatfield, K.~Simonyan, A.~Vedaldi, and A.~Zisserman, ``Return of the devil
  in the details: Delving deep into convolutional nets,'' \emph{arXiv preprint
  arXiv:1405.3531}, 2014.

\bibitem{VGG-VDNets}
K.~Simonyan and A.~Zisserman, ``Very deep convolutional networks for
  large-scale image recognition,'' in \emph{International Conference on
  Learning Representations}, 2015.

\bibitem{GoogLeNet}
C.~Szegedy, W.~Liu, Y.~Jia, P.~Sermanet, S.~Reed, D.~Anguelov, D.~Erhan,
  V.~Vanhoucke, and A.~Rabinovich, ``Going deeper with convolutions,'' in
  \emph{Proceedings of the IEEE Conference on Computer Vision and Pattern
  Recognition}, 2015, pp. 1--9.

\bibitem{Hybrid}
A.~Mousavian and J.~Kosecka, ``Deep convolutional features for image based
  retrieval and scene categorization,'' \emph{arXiv preprint arXiv:1509.06033},
  2015.

\bibitem{SPoC}
A.~Babenko and V.~Lempitsky, ``Aggregating deep convolutional features for
  image retrieval,'' \emph{arXiv preprint arXiv:1510.07493}, 2015.

\bibitem{CroW}
Y.~Kalantidis, C.~Mellina, and S.~Osindero, ``Cross-dimensional weighting for
  aggregated deep convolutional features,'' in \emph{European Conference on
  Computer Vision 2016 Workshops}, 2016.

\bibitem{RSSCN7}
Q.~Zou, L.~Ni, T.~Zhang, and Q.~Wang, ``Deep learning based feature selection
  for remote sensing scene classification,'' \emph{Geoscience and Remote
  Sensing Letters, IEEE}, vol.~12, no.~11, pp. 2321--2325, 2015.

\bibitem{UCM}
Y.~Yang and S.~Newsam, ``Bag-of-visual-words and spatial extensions for
  land-use classification,'' in \emph{Proceedings of the 18th SIGSPATIAL
  International Conference on Advances in Geographic Information Systems},
  2010, pp. 270--279.

\bibitem{ANMRR}
B.~S. Manjunath, J.-R. Ohm, V.~V. Vasudevan, and A.~Yamada, ``Color and texture
  descriptors,'' \emph{Circuits and Systems for Video Technology, IEEE
  Transactions on}, vol.~11, no.~6, pp. 703--715, 2001.

\end{thebibliography}

\end{document}